\documentclass[sigconf]{acmart}

\AtBeginDocument{%
  }

\acmYear{2022}\copyrightyear{2022}
\setcopyright{acmlicensed}
\acmConference[MobiSys '22]{The 20th Annual International Conference on Mobile Systems, Applications and Services}{June 25--July 1, 2022}{Portland, OR, USA}
\acmBooktitle{The 20th Annual International Conference on Mobile Systems, Applications and Services (MobiSys '22), June 25--July 1, 2022, Portland, OR, USA}
\acmPrice{}
\acmDOI{10.1145/3498361.3538917}
\acmISBN{978-1-4503-9185-6/22/06}

\usepackage{amsmath,amsfonts,bm}

\def\eqref#1{equation~\ref{#1}}

\def\1{\bm{1}}

\def\evw{{w}}

\DeclareMathAlphabet{\mathsfit}{\encodingdefault}{\sfdefault}{m}{sl}
\SetMathAlphabet{\mathsfit}{bold}{\encodingdefault}{\sfdefault}{bx}{n}

\usepackage{enumitem}
\usepackage{booktabs,siunitx}
\usepackage{caption}
\usepackage{subcaption}
\usepackage{algorithm}
\usepackage{ulem}
\usepackage[noend]{algpseudocode}
\usepackage{multirow}

\newcommand{\rev}[1]{\textcolor{black}{#1}}

\begin{document}


\title{FedBalancer: Data and Pace Control for Efficient Federated Learning on Heterogeneous Clients}


\author{Jaemin Shin}
\affiliation{%
 \institution{School of Computing\\KAIST}
 \city{Daejeon}
 \country{Republic of Korea}}
\email{jaemin.shin@kaist.ac.kr}

\author{Yuanchun Li}
\authornote{corresponding author.}
\affiliation{%
  \institution{Institute for AI Industry Research (AIR)\\Tsinghua University}
 \city{Beijing}
 \country{China}}
\email{liyuanchun@air.tsinghua.edu.cn}

\author{Yunxin Liu}
\affiliation{%
  \institution{Institute for AI Industry Research (AIR)\\Tsinghua University}
 \city{Beijing}
 \country{China}}
\email{liuyunxin@air.tsinghua.edu.cn}

\author{Sung-Ju Lee}
\affiliation{
  \institution{School of Electrical Engineering\\KAIST}
  \city{Daejeon}
  \country{Republic of Korea}}
\email{profsj@kaist.ac.kr}

\renewcommand{\shortauthors}{Jaemin Shin, et al.}

\makeatletter
\let\@authorsaddresses\@empty
\makeatother

\begin{abstract}

Federated Learning (FL) trains a machine learning model on distributed clients without exposing individual data. 
Unlike centralized training that is usually based on carefully-organized data, FL deals with on-device data that are often unfiltered and imbalanced.  As a result, conventional FL training protocol that treats all data equally leads to a waste of local computational resources and slows down the global learning process.  
To this end, we propose \putname{}, a systematic FL framework that actively selects clients' training samples. Our sample selection strategy prioritizes more ``informative'' data while respecting privacy and computational capabilities of clients. To better utilize the sample selection to speed up global training, we further introduce an adaptive deadline control scheme that predicts the optimal deadline for each round with varying client training data. 
Compared with existing FL algorithms with deadline configuration methods, our evaluation on five datasets from three different domains shows that \putname{} improves 
the time-to-accuracy performance by \rev{1.20$\sim$4.48$\times$} 
while improving the model accuracy by \rev{1.1$\sim$5.0\%}. We also show that \putname{} is readily applicable to other FL approaches by demonstrating that \putname{} improves the convergence speed and accuracy 
when operating jointly with three different FL algorithms.






\end{abstract}



\newcommand{\putname}{\textit{FedBalancer}}

\renewcommand{\algorithmicrequire}{\textbf{Input:}}
\renewcommand{\algorithmicensure}{\textbf{Output:}}
\renewcommand{\algorithmicprocedure}{\textbf{func}}

\maketitle

\section{Introduction}

Federated learning (FL) is a machine learning paradigm that performs decentralized training of models on mobile devices (e.g., smartphones) with locally stored data~\cite{mcmahan01}. FL trains on a large corpus of private user data without collecting them, with only the model weight updates communicated externally from the user’s device~\cite{bonawitz01}. With FL, researchers have proposed to improve AI in diverse domains: human mobility prediction~\cite{feng01}, RF localization~\cite{ciftler01}, traffic sign classification~\cite{albaseer01}, tumour detection~\cite{li02, jimenez01}, and Clinical Decision Support (CDS) model for COVID-19~\cite{dayan01}. FL also has product deployments as large companies such as Google or Taobao deploy language processing and item recommendation tasks across millions of real-world devices~\cite{niu01, yang02}.

One of the key objectives in FL is to optimize \textit{time-to-accuracy} performance, which is a wall clock time for a model to achieve the target accuracy~\cite{lai01}. Achieving high time-to-accuracy performance is important as FL consumes significant computation and network resources on edge-user devices~\cite{dinh01}. For model developers who prototype a mobile AI with FL without a proxy dataset, achieving faster convergence on thousands to millions of devices is desired to efficiently test multiple model architectures and hyperparameters~\cite{kairouz01}. Service providers who frequently update a model with continual learning with FL require to minimize the user overhead with better time-to-accuracy performance~\cite{le01}. 


In realistic FL scenarios, however, heterogeneous data being distributed over isolated users becomes the main challenge in achieving high time-to-accuracy performance~\cite{zhao01, li03}. While data engineering techniques such as 
\textit{importance sampling}~\cite{katharopoulos01, alain01, loshchilov01, schaul01} are widely adopted in centralized learning to optimize the training process, applying them in FL is infeasible as it requires private user data sharing. For this reason, previous FL algorithms~\cite{mcmahan01, li01, lai01} mostly treat every client data equally, which could result in a waste of computational resources and slow convergence. We conducted a preliminary study to examine this phenomenon and found that the ratio of informative samples is reduced from 93.2\% to 20\% as the FL progresses.
This could further exacerbate with heterogeneous hardware of FL clients, as low-end devices might fail to send model updates while training a large portion of unimportant samples.


To address this problem, we propose \putname{}, a systematic FL framework with sample selection. The sample selection of \putname{} prioritizes more ``informative'' samples of clients to efficiently utilize their computational effort. This allows low-end devices to contribute to the global training within the round deadline by focusing on smaller but more important training samples. To achieve high time-to-accuracy performance, the sample selection is designed to operate without additional forward or backward pass for sample utility measurement at FL rounds. Lastly, \putname{} can coexist and collaborate with orthogonal FL approaches to further improve performance. 

To realize \putname{}, we addressed the following challenges: (1) Simply reducing the training data of a client with random sampling could lead to lower model accuracy as the statistical utility of the training data would decrease. As such, \putname{} selects samples based on their statistical utility measurement. (2) Collecting sample-level statistical utility for sample selection might break the privacy guarantee of FL. To address this problem, we propose client-server coordination to maintain \textit{loss threshold}, which allows clients to effectively select important samples while only exposing differentially-private statistics of their data. (3) 
The sample selection itself might not directly lead to time-to-accuracy performance improvement when the FL round deadline remains fixed. 
To formulate the benefit of selecting different deadlines, we propose a metric \textit{deadline efficiency} (DDL-E) 
that calculates the number of round-completed clients per time. This allows \putname{} to predict the optimal deadline with varying client training data.

We implemented \putname{} 
and conducted experiments on five datasets from three domains that contain real-world user data. Compared with existing FL aggregation algorithms with deadline configuration methods, \putname{} improves the time-to-accuracy performance by 1.20$\sim$4.48$\times$. \putname{} achieves this improvement without sacrificing model accuracy; in fact, it improves the accuracy by 1.1\%$\sim$5.0\%. In addition, we implement \putname{} on top of three orthogonal FL algorithms to demonstrate that \putname{} is easily applicable to other FL approaches and improves their time-to-accuracy performance. 

We summarize our contributions as follows:

\begin{itemize}[topsep=0pt, align=left, labelwidth=10pt, leftmargin=15pt]
\item We propose a systematic framework for FL with sample selection, which actively selects high utility samples at each round without collecting privacy-invasive sample-level information from clients.
\item We propose a deadline control strategy for each round of FL based on the newly defined metric \textit{deadline efficiency}, which optimizes the time-to-accuracy performance along with our sample selection.
\item We implement \putname{} jointly with existing FL algorithms, showing improvement in both time-to-accuracy and model accuracy.
\end{itemize}












\section{Background and Motivation}

\subsection{Federated Learning}
\label{sec:fedlearning}



Federated Learning (FL) operates across multiple mobile devices to globally train a model from the distributed client data. FedAvg~\cite{mcmahan01}, the most commonly used FL approach~\cite{jiang01}, operates as follows: (i) Suppose there are $N$ clients in an FL system. For each round of FL, the server randomly selects $K$ clients ($K<<N$) who participate in model training. (ii) At the \textit{R}-th round, the server transmits the current model weights $\displaystyle \evw_R$ to the selected clients. (iii) Each client then performs model training for $E$ epochs with their local data and generates $\displaystyle \evw^k_{R+1}$, where $k$ denotes the client index. (iv) Clients upload the updated model parameters to the server, and (v) the server aggregates different clients' updates and generate the updated model as $\evw_{R+1} \leftarrow \sum^K_{k=1} {\frac{n_k}{n}}{\evw^k_{R+1}}$,
where $n_k$ indicates the number of data points of client $k$ and $n$ the number of all data points.





A key objective in FL is to optimize time-to-accuracy performance. FL tasks typically require hundreds to thousands of rounds to converge~\cite{caldas01, lai01}, and clients participating at a round undergo substantial computational and network overhead~\cite{dinh01}. Deploying FL across thousands to millions of devices should be done efficiently, quickly reaching the model convergence while not sacrificing the model accuracy. This becomes more important when FL has to be done multiple times, as often the case when model developers prototype a new model with FL without a proxy dataset or periodically update a deployed model to new domain via continual learning or online learning with FL.


However, the heterogeneities of the real-world mobile clients make it challenging to achieve good time-to-accuracy performance. \newline \noindent \textbf{Data heterogeneity}: The client-generated training data are imbalanced and not independent and identically distributed (non-IID) due to each user's different mobile device usage or physical and mental characteristics~\cite{mcmahan01, wu01}. While the training data in centralized learning could be filtered and organized with data engineering techniques~\cite{katharopoulos01, alain01, loshchilov01, schaul01} to address data heterogeneity, applying the same solution is infeasible in FL as it is dealing with distributed 
private data on clients. Such data heterogeneity of FL clients results in slow convergence and suboptimal performance~\cite{li01, zhao01, li03}. \newline \noindent \textbf{Hardware heterogeneity}: The client devices 
have distinct computational capabilities and network connectivity, resulting in up to 12$\times$ more round completion time between different clients~\cite{li04}. 
Thus, waiting for every client to complete its task at a round might significantly delay the training process. A common practice for such an issue is to set a \textit{deadline} for a round duration and drop clients that fail to send the model updates before the deadline~\cite{nishio01, abdelmoniem01}. However, this results in less contribution from clients with low-end devices, which could result in delayed convergence or biased model training~\cite{yang01, li01}.

\subsection{Motivational Study}
\label{sec:motivation}

To address the heterogeneity problems, various approaches have been proposed. Researchers proposed FL algorithms that allow clients to train different numbers of local epochs~\cite{li01} or different subsets of model weights~\cite{diao01, horvath01} based on clients' hardware capabilities. Model personalization has also been proposed to tackle data heterogeneity~\cite{pillutla01, ouyang01, fallah01, tu01, li06, li07, liu2020pmc, liu2021distfl}. In addition, client selection strategies for FL training have been proposed to optimize the convergence speed on heterogeneous clients~\cite{lai01, cho01, cho02}. Although these techniques have improved the convergence speed or accuracy, they treat all data of clients equally, which could lead to a waste of computational resources for training non-important samples and result in suboptimal time-to-accuracy performance. 


We investigate such limitation of previous FL approaches and discuss how \putname{} should be designed based on our experiments that simulate FL on heterogeneous clients. We used widely used datasets to benchmark FL methods: FEMNIST~\cite{cohen01} and Shakespeare~\cite{shakespeare01}. 
We simulated heterogeneous training latency and network connectivity on the real-world clients as described in Section~\ref{sec:experimentalsetup}.

\begin{figure}[t]
    \centering
    \begin{subfigure}[b]{0.235\textwidth}
        \centering
        \includegraphics[width=\textwidth]{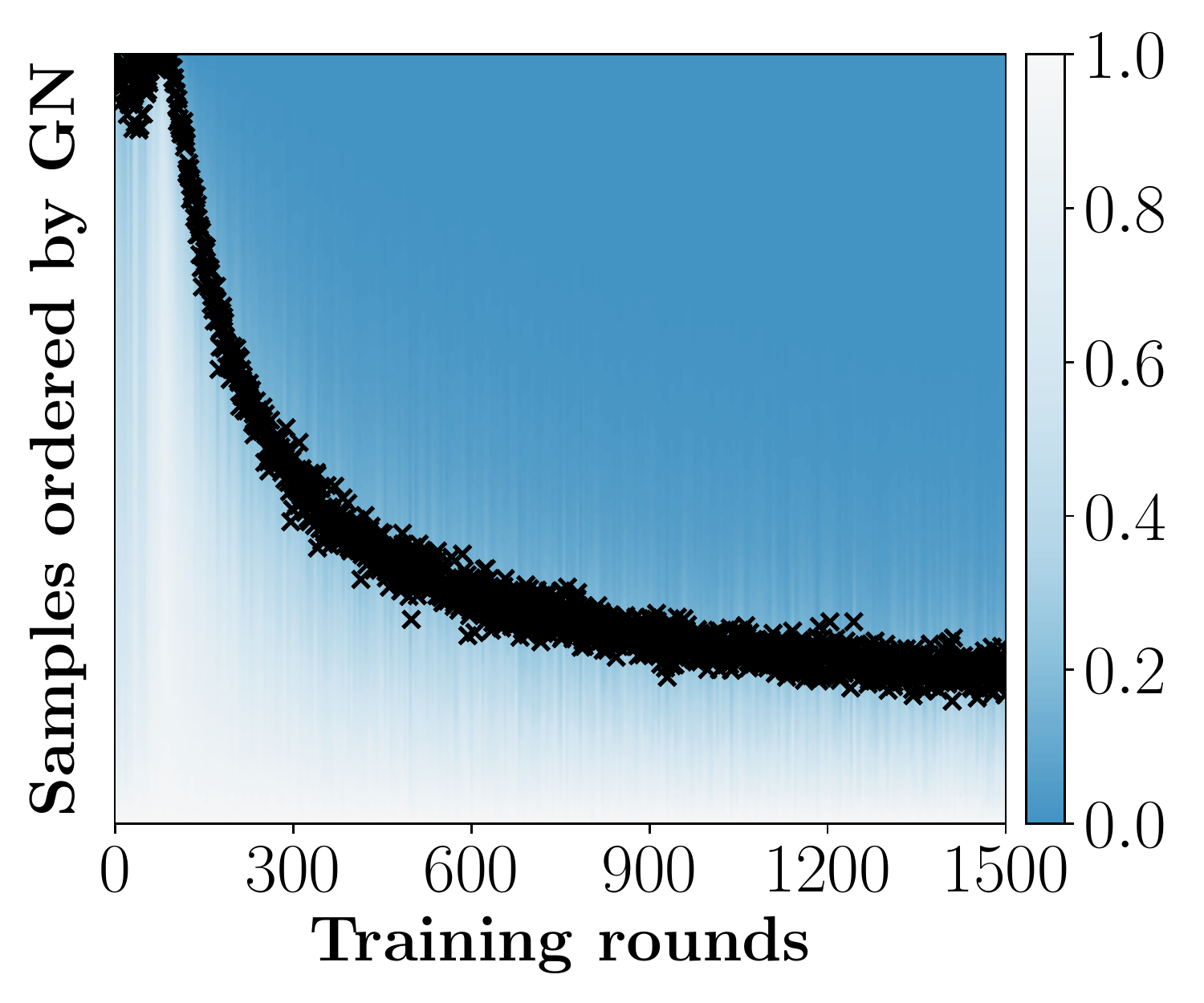}
        \caption{FEMNIST dataset.}
        \label{fig:motivation-1}
    \end{subfigure}
    \hfill
    \begin{subfigure}[b]{0.235\textwidth}
        \centering
        \includegraphics[width=\textwidth]{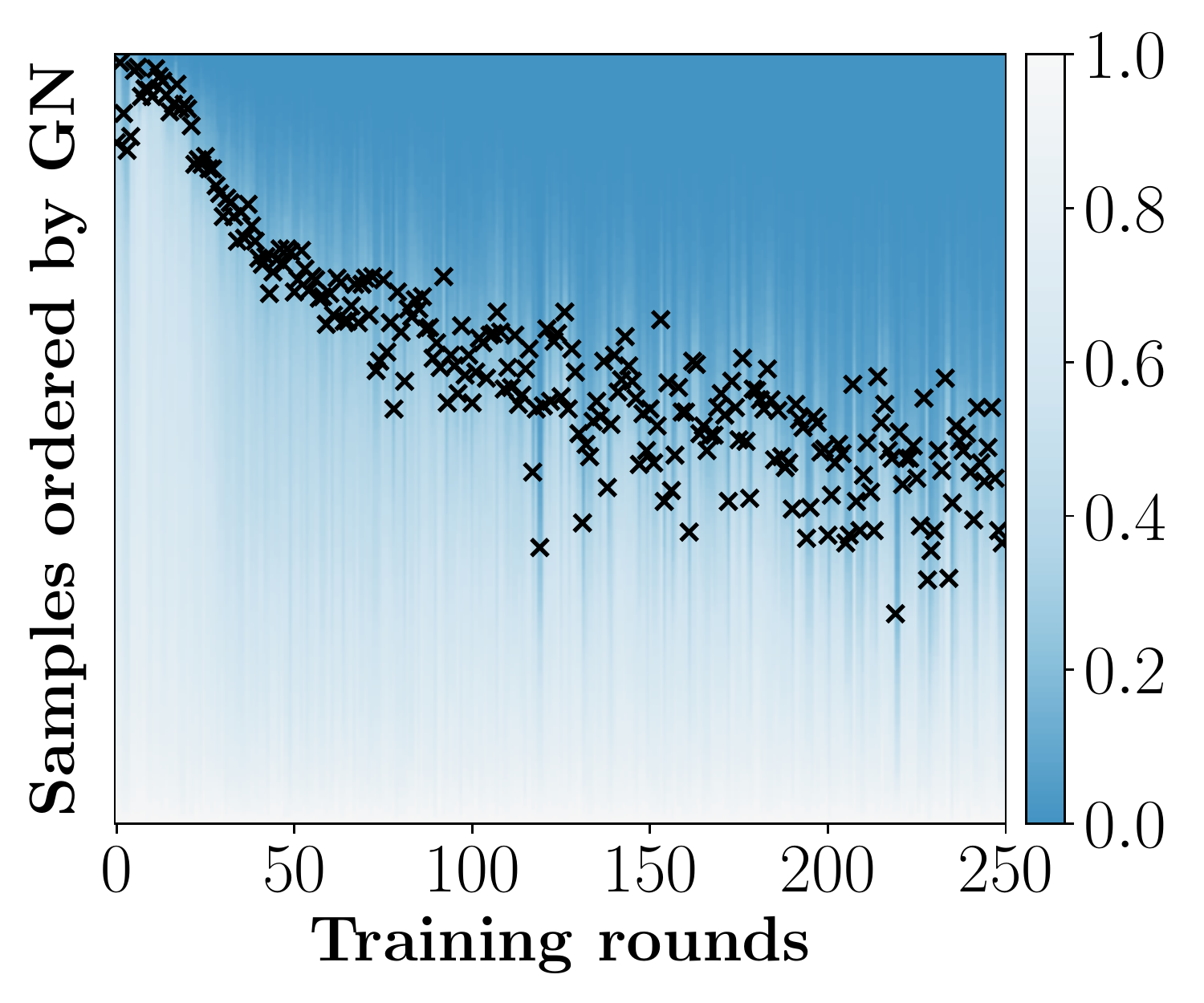}
        \caption{Shakespeare dataset.}
        \label{fig:motivation-2}
    \end{subfigure}
  \caption{Ordered gradient norm (GN) of samples from FL training rounds on two different datasets.
  }
  \vspace{-0.2cm}
  \label{fig:motivation-gn}
  \vspace{-0.2cm}
\end{figure}

\textbf{Inefficiency of Full Data Training.} As in FedAvg~\cite{mcmahan01}, most FL approaches let clients to fully train their entire data at a training round. Other approaches that samples a subset of client training data for each round use equal weights on all data~\cite{lai01, fallah01, jiang02}. However, previous studies in centralized machine learning~\cite{katharopoulos01, alain01, loshchilov01} found that the importance of all samples are not equal; A large portion of samples are learned quickly after few training rounds and could be ignored afterwards. Thus, we conducted an experiment to verify if the same phenomenon also applies in FL and how significant it is. We suspect that the inefficiency problem would be more serious in FL as limited computational resources of mobile clients could be wasted on non-important samples.

In this experiment, we measured the Gradient Norm (GN) of each sample on federated clients to evaluate sample-level contribution on a model update. For each training round, we collected and sorted the GN of each sample from the selected clients. We removed top 5\% of samples to avoid evaluating noisy samples and applied min-max scaling to [0.0, 1.0] interval. The experiment ran until the model converged. The results from each dataset are shown in Figure~\ref{fig:motivation-1} and~\ref{fig:motivation-2}. Each column of graphs indicates the sorted GN of samples from a training round, where the largest value is located at the bottom. The x-shaped points illustrate where the gradient norm with scaled value of 0.2 exists at each training round. 

From both datasets, we observe that the GN of samples are mostly high at the early stage of FL, but only small portion of samples show high GN afterwards, meaning that the number of samples that contribute knowledge to the model are reducing as the training progresses. In FEMNIST dataset, only 6.8\% of the samples from early training rounds (round index 0$\sim$150) have less scaled GN value than 0.2, but 80.0\% of the samples are less than 0.2 for the last 150 training rounds. Similarly in the Shakespeare dataset, 6.7\% of the samples from early training rounds (round index 0$\sim$25) have less scaled GN value than 0.2, but 54.8\% of the samples are less than 0.2 for the last 25 
training rounds. This result also implies that the samples are not equally important during these FL tasks and current FL approaches could spend large portion of training time for samples that have negligible contribution to the model update. 

This experiment motivates \putname{} to start training with all the samples at the beginning, and then gradually remove samples that the model has already learned. In Section~\ref{sec:cds}, we further illustrate how \putname{} selects a subset of samples of each client at training rounds based on our findings in this experiment. 



\begin{figure}[t]
    \centering
    \begin{subfigure}[b]{0.235\textwidth}
        \centering
        \includegraphics[width=\textwidth]{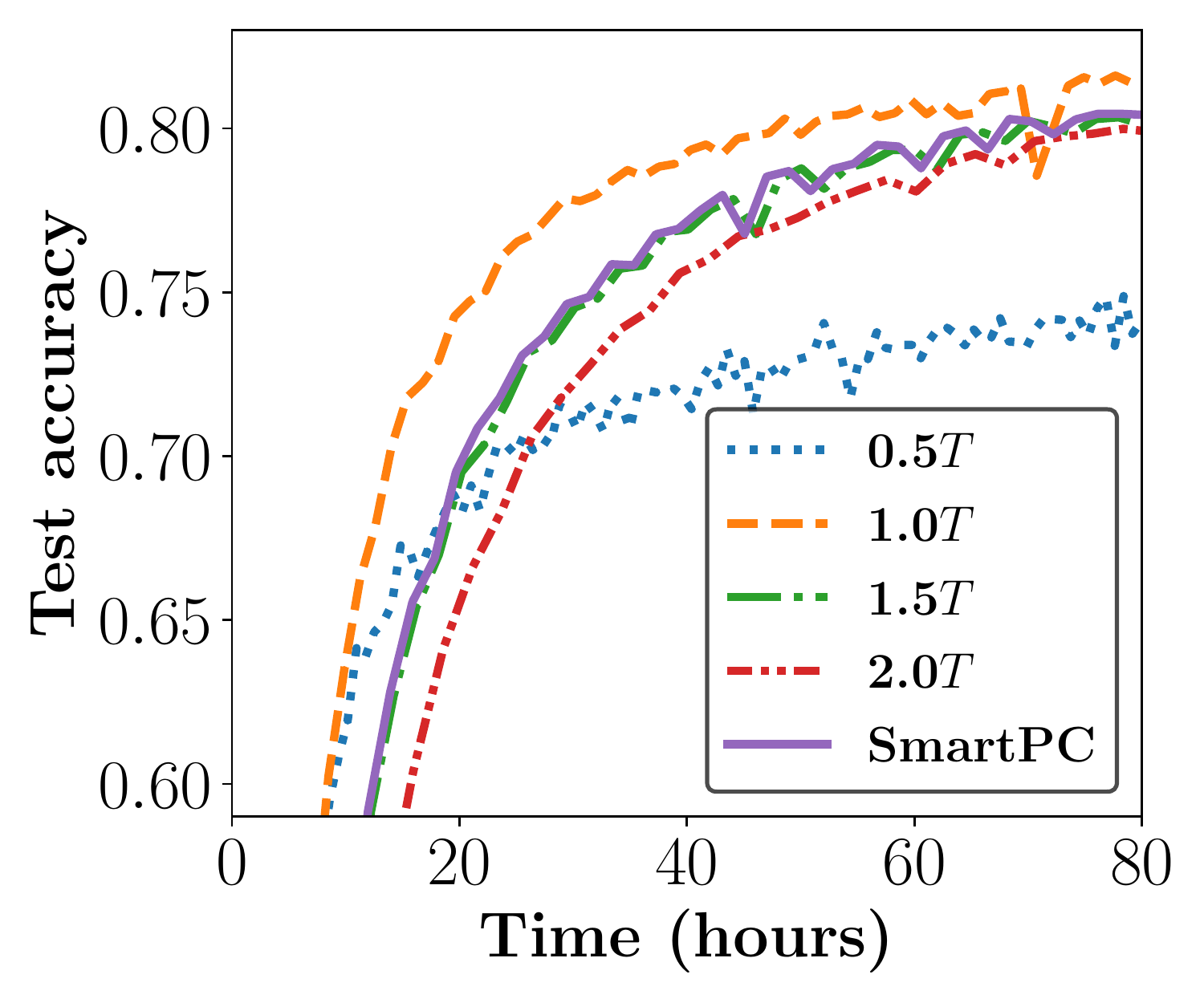}
        \caption{FEMNIST dataset.}
        \label{fig:motivation-3}
    \end{subfigure}
    \hfill
    \begin{subfigure}[b]{0.235\textwidth}
        \centering
        \includegraphics[width=\textwidth]{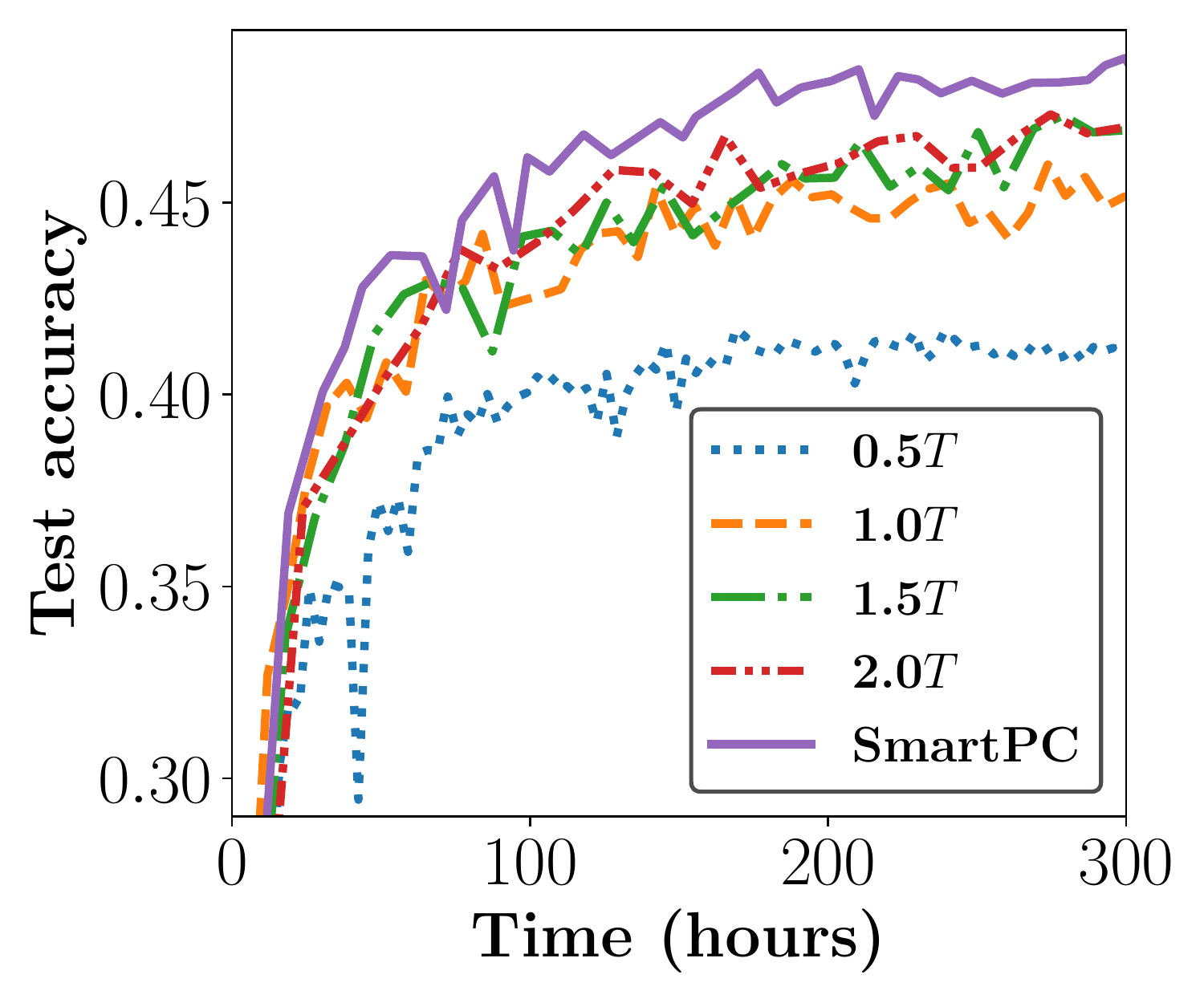}
        \caption{Shakespeare dataset.}
        \label{fig:motivation-4}
    \end{subfigure}
  \caption{FL on two datasets with different deadline configuration methods.
  }
  \vspace{-0.5cm}
  \vspace{-0.5cm}
  \label{fig:motivation-ddl}
\end{figure}

\begin{figure*}[t]
  \centering
  \includegraphics[width=0.9\linewidth]{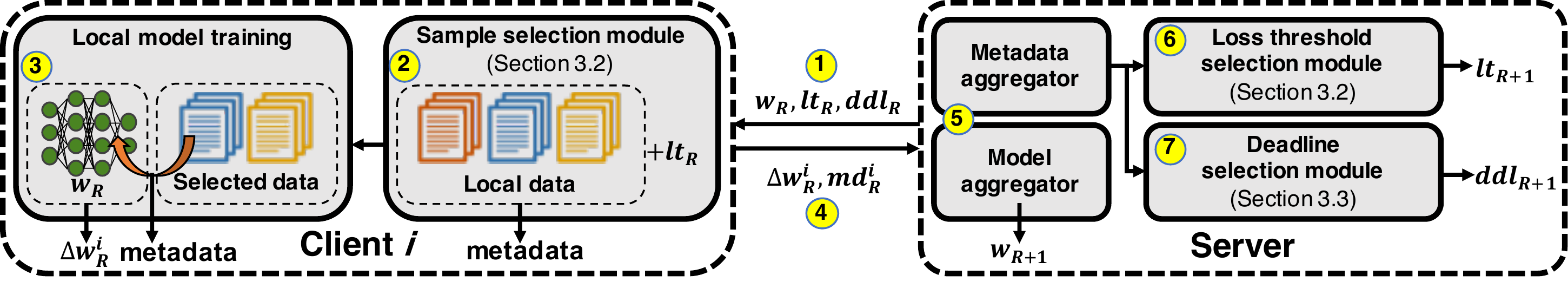}
  \vspace{-0.1cm}
  \caption{Overview of \putname{} architecture and its operation in an FL round in seven steps. 
  }
  \label{fig:overview}
  \vspace{-0.1cm}
\end{figure*}

\textbf{Importance of Deadline Selection.} While the existence of \textit{optimal deadline} for achieving shortest training time in FL has been studied~\cite{yang01}, controlling the deadline for high time-to-accuracy performance has been largely overlooked. 
To understand the performance of existing deadline configuration methods, we conducted an experiment on the two datasets 
with SmartPC~\cite{li04} and four different fixed deadlines --- 0.5$T$, 1.0$T$, 1.5$T$, and 2.0$T$ --- where $T$ indicates the mean of round completion time on every participating client. For SmartPC, we implemented a training round to last until 80\% of the clients complete their task, where 80\% is suggested by Li et al.~\cite{li04}.

Figure~\ref{fig:motivation-3} and ~\ref{fig:motivation-4} illustrate the results on two datasets. Our takeaway from this experiment is twofold: (1) The deadline is a significant factor in achieving fast convergence speed and high accuracy, and (2) no single method achieved the best performance for both FL tasks.
In FEMNIST dataset, deadline 1.0$T$ achieved the highest final accuracy (.815), while being 43.6\%, 47.0\% and 77.9\% faster than SmartPC, deadline 1.5$T$, and deadline 2.0$T$, respectively, in achieving the test accuracy of .750. On the other hand, in Shakespeare dataset, SmartPC achieved the  highest final accuracy (.488) while being 61.7\%, 42.8\%, and 45.4\% faster than deadline 1.0$T$, 1.5$T$, and 2.0$T$, respectively, in achieving the test accuracy of .450. Deadline 0.5$T$ could not achieve high accuracy in either task, as most clients failed to upload their model update within the deadline.



In a training round of FL, the computation time has been shown to be the bottleneck~\cite{wang02, pilla01, yang01}. As the 
amount of client computation changes with the sample selection of \putname{}, finding an optimal deadline could be an important problem in achieving high time-to-accuracy performance. To this end, in Section~\ref{sec:adc}, we propose how \putname{} finds the optimal deadline for each training round for efficient FL on heterogeneous clients. 




\section{\putname{}}

\subsection{Overview}
\label{sec:overview}


For each round of FL, \putname{} adaptively selects the client training data and controls the deadline to achieve high time-to-accuracy performance. We first provide an overview of how \putname{} operates in an FL round and then describe how each component of \putname{} is designed.

Figure~\ref{fig:overview} depicts the \putname{} architecture during an FL round. The main functionality of \putname{} is to actively control two variables for each round: \textit{loss threshold} ($lt$) and \textit{deadline} ($ddl$). The loss threshold works as a parameter that determines each client's training data (Section~\ref{sec:cds}) and the \textit{deadline} determines the round termination time. The numbers inside a circle show the seven steps of an FL round with \putname{}.

The server first transmits the current model weights $\displaystyle \evw_R$, the loss threshold $lt_R$, and the deadline $ddl_R$ (\textit{R} indicates the \textit{R}-th round) to the selected clients of the round (Step 1). The \textit{sample selection module} at each client selects the partial training data with the received loss threshold (Step 2) and trains the received model (Step 3). The client transmits the model update and the \textit{metadata} collected from the sample selection and model training (Step 4), and the server aggregates these responses from all clients (Step 5). Based on the metadata from clients, the \textit{loss threshold selection module} and the \textit{deadline selection module} each selects the loss threshold $lt_{R+1}$ and the deadline $ddl_{R+1}$ for the next round (Steps 6 and 7).

\textbf{Challenges: } We aim to address the following challenges to realize \putname{}:


\begin{enumerate}[topsep=3pt, align=left, labelwidth=10pt, leftmargin=15pt]
  \item \textit{Sample selection without accuracy drop:} 
  Simply reducing the training data with random sampling could result in degradation of model accuracy due to the decreased statistical utility. \putname{} should thus prioritize samples based on their statistical utility.
  \item \textit{Privacy-preserving sample selection in FL}: While we aim to select an optimal set of client training data for each round, requiring up-to-date sample-level information from clients could harm the privacy guarantee of FL. \putname{} should select client training samples without collecting privacy-invasive information from clients.
  \item \textit{Predicting optimal deadline with varying data}: As computation could be the bottleneck of an FL round~\cite{wang01, pilla01, yang01}, applying sample selection strategy of \putname{} might greatly change the round completion time of clients. \putname{} should  adaptively predict the optimal deadline based on the sample selection status of heterogeneous clients for each round of FL.
\end{enumerate}

\subsection{Client Sample Selection}
\label{sec:cds}


In Section~\ref{sec:motivation}, we observed that existing FL methods consume large portion of time to train samples that contribute only small gradient to the model. As these samples are quickly learned after few rounds, we design \putname{} to start training with all samples and gradually remove \textit{already-learned} samples. This enables \putname{} to efficiently focus on more \textit{important} samples at each round while optimizing the training process of FL. However, implementing such design in FL is non-trivial as the following question needs to be addressed: \textit{How should \putname{} distinguish between \textit{important} and \textit{non-important} samples at each stage of FL?}

A straightforward solution is to collect every sample-level importance information from all clients to a server at each round, and derive a criteria that determines more important samples to a current model. However, such approach is hardly applicable in FL as sharing information of every sample could break the privacy guarantee of FL and reveal the clients' data. This approach also incurs significant network overhead in communicating all sample's information at each round. An alternative approach is to have clients classify more important samples within their local data without exposing any information. However, as client data distributions are heterogeneous in FL~\cite{mcmahan01, wu01, li01, zhao01, li03}, clients could struggle to determine important samples without knowing the global data distribution. 

To address this issue, we propose a client-server coordination to maintain a \textit{loss threshold} variable, which enables clients to effectively select important samples without exposing private sample-level information. 
\putname{} actively controls the loss threshold based on the collected metadata from clients with \textit{loss threshold selection module}, where the metadata consists of differentially-private statistics of sample-level information. 

Note that \putname{} uses the \textit{loss} of a sample to measure the statistical utility (and thus the importance) of a sample to the current model, 
similar to \textit{Importance Sampling}~\cite{loshchilov01, schaul01}. While other studies have also leveraged \textit{gradient norm} or \textit{gradient norm upper bound}~\cite{katharopoulos01, alain01, li08} to achieve the same goal, we use loss as it is more widely applicable to FL tasks with non gradient-based optimizations~\cite{rios01}.

\subsubsection{Sample selection module}

\begin{algorithm}[t]
\caption{Client $i$ at $R$-th round: Sample selection 
}\label{alg:ssm}

\begin{algorithmic}[1]
\Procedure{SelectSample}{$D^i$, $B^i$, $loss^i$, $lt_{R}$, $ddl_{R}$, $\displaystyle \evw_{R}$, $p$, $E$}
\State $S \gets $\hspace{0.1cm}\texttt{maxTrainableSize}$(mean(B^i), ddl_{R}, E)$ \label{algline:ssm-line2}
\If{$S \geq len(D^i)$} \label{algline:ssm-line3}
\State $D^i_{R} \gets D^i$ \label{algline:ssm-line4}
\Else
\State $D^i_{R}, OT^i, UT^i \gets \emptyset$
\For{$x \gets 1$ to $len(D^i)$} \label{algline:ssm-line8}
    \State \textbf{if} $loss^i[x] \geq \displaystyle lt_{R}$ \textbf{then} $OT^i.insert(D^i[x])$ \label{algline:ssm-line9}
    \State \textbf{else} $UT^i.insert(D^i[x])$ \label{algline:ssm-line10}
\EndFor \label{algline:ssm-line11}
\State $L \gets max(S, len(OT^i))$ \label{algline:ssm-linenew}
\State $D' \gets $\hspace{0.1cm}\texttt{randSample}$(OT^i, L \cdot p))$ \label{algline:ssm-line16}
\State $D'' \gets $\hspace{0.1cm}\texttt{randSample}$(UT^i, L \cdot (1 - p))$\label{algline:ssm-line17}
\EndIf \label{algline:ssm-line18}
\State $D^i_{R} \gets $\texttt{concatenate}$(D', D'')$
\State \textbf{return} $D^i_{R}$ 
\EndProcedure
\end{algorithmic}
\end{algorithm}


Algorithm~\ref{alg:ssm} describes how the \textit{sample selection module} of a client selects the samples at each round after it receives the loss threshold from the server. Let us assume that client $i$'s sample selection module is working at round $R$, with a given loss threshold $lt_R$. 

First, the module measures if a sample selection on a client is required for this round --- i.e., it measures if a client is fast enough to train its full dataset within the given deadline $ddl_R$. While \putname{} makes clients focus on more important samples for efficient training, it allows clients to fully utilize its statistical utility if feasible. To this end, we calculate the maximum number of samples $S$ which client $i$ can train for $E$ epochs before the deadline, and verify if it is larger than the size of the client dataset $D^i$ (Line~\ref{algline:ssm-line2} -~\ref{algline:ssm-line4}). As the computational ability of a client can change according to the device runtime conditions~\cite{yang03}, \putname{} collects the batch training latency of a client as $B^i$ during FL and uses the mean latency to estimate the max samples it can process. To calculate the mean latency from the first round, \putname{} asks clients to sample the latency for $k$ times before FL begins. We used 10 for $k$ in our evaluation.


If the client $i$ is incapable of training its full dataset, the \textit{sample selection module} determines which samples to train at the FL round by using the \textit{list of sample loss} ($loss^i$). The loss list shows the statistical utility of all samples on the current model. While such loss list could be obtained by inferring all samples on up-to-date model at each round, it requires additional forward pass latency that could degrade the time-to-accuracy performance. Therefore, clients of \putname{} perform whole-dataset forward pass only once when they are first selected at a round to generate a loss list. Then, whenever they train the subset of data, they update the loss values of selected samples that are obtained from the training procedure. We discuss the trade-off between latency reduction and obtaining up-to-date information in Section~\ref{sec:latencytradeoff}.



\putname{} selects client samples based on the list of sample loss as follows: First, it divides client i's samples into two groups: Under-Threshold ($UT^i$) and Over-Threshold ($OT^i$). Samples that have smaller loss than the loss threshold $lt_R$ are put in $UT^i$ and otherwise in $OT^i$ (Line~\ref{algline:ssm-line8} -~\ref{algline:ssm-line11}). We regard samples in $OT^i$ to be more \textit{important} samples in training the current model and prioritize them in sample selection. We sample $L\cdot p$ samples from $OT^i$ and $L\cdot(1 - p)$ samples from $UT^i$ where $L$ indicates the number of selected samples and $p$ is a parameter in an interval of [0.5, 1.0] (Line~\ref{algline:ssm-linenew} -~\ref{algline:ssm-line18}). The intuition of sampling a portion of data from $UT^i$ is to avoid \textit{catastrophic forgetting}~\cite{kirkpatrick01, yoon01} of the model on \textit{already-learned} samples.

The number of selected samples, $L$, is determined as the number of samples in $OT^i$ ($len(OT^i)$). The loss threshold gradually increases (explained in Section~\ref{sec:ltsm}), which allows clients to efficiently focus on samples with high statistical utility. However, if $S$ is larger than $len(OT^i)$, a client instead uses $S$ 
to maximize the statistical utility within the deadline (Line~\ref{algline:ssm-linenew}). As \putname{} is built on top of Prox~\cite{li01} that allows clients to train less number of epochs, clients with $S$ less than $len(OT^i)$ could still contribute to the model update. 


\subsubsection{Loss threshold selection module}
\label{sec:ltsm}

\begin{algorithm}[t]
\caption{$lt$ selection for next ($R+1$)-th round}\label{alg:ps}
\begin{algorithmic}[1]
\Procedure{SelectLossThreshold}{$LLow_{R}$, $LHigh_{R}$, $ltr$} \label{algline:ps-line1}
\State $ll \gets min(LLow_{R})$ \label{algline:ps-line2}
\State $lh \gets mean(LHigh_{R})$ \label{algline:ps-line3}
\State $lt_{R+1} \gets ll + (lh - ll) \cdot ltr$ \label{algline:ps-line4}
\State \textbf{return} $lt_{R+1}$ \label{algline:ps-line5}
\EndProcedure\label{algline:ps-line6}
\end{algorithmic}
\end{algorithm}


The \textit{loss threshold selection module} determines a loss threshold that effectively distinguishes the \textit{important} and \textit{non-important} samples. As the loss distribution of samples changes as FL proceeds, it is essential for the module to be knowledgeable of the current distribution. To respect privacy, the server collects few statistical information from the loss list of clients as a metadata at the end of each round. 
Specifically, client $i$ at the $R$-th round provides $LLow^i_R$ and $LHigh^i_R$ values to the server, which indicate the low and high loss value of its current samples, respectively. We use the min loss value as $LLow^i_R$ and use 80\% percentile loss 
from the list as $LHigh^i_R$ instead of the max value, as noisy samples could have abnormally high loss~\cite{song01} and make \putname{} to misjudge the loss distribution. As such values directly indicate a loss value of a specific sample, we apply Gaussian noise to the values to protect user privacy, as in differential privacy~\cite{lai01, mcmahan02, wei01}. These values from clients get further aggregated on the server into a list as $LLow_R$ and $LHigh_R$. We report the performance of \putname{} when different levels of noise is applied in  Section~\ref{sec:parametersensitivityanalysis}. 

With these metadata, the loss threshold selection module selects a loss threshold as described in Lines~\ref{algline:ps-line1} -~\ref{algline:ps-line6} in Algorithm~\ref{alg:ps}. The module measures the \textit{loss low} value ($ll$) and \textit{loss high} value ($lh$) to estimate the current range of sample loss values of the clients (Lines~\ref{algline:ps-line2} and ~\ref{algline:ps-line3}). The module then outputs the loss threshold of the next round ($R+1$) with variable \textit{loss threshold ratio} ($ltr$), which calculates the linear interpolation between $ll$ and $lh$ (Line~\ref{algline:ps-line4}) as  $lt_{R+1} \leftarrow ll + (lh - ll) \cdot ltr$. 
The loss threshold ratio $(ltr)$ enables \putname{} to start training with all samples and gradually remove \textit{already-learned} samples. \putname{} initialize $ltr$ as 0.0 and gradually increases the value by \textit{loss threshold step size} $(lss)$ as shown in Algorithm~\ref{alg:rc}. Note that the \textit{deadline ratio} $(ddlr)$, which controls the deadline of each round (described in Section~\ref{sec:adc}), is also controlled with $ltr$. 

To control $ltr$ and $ddlr$, \putname{} evaluates the benefit of the current configuration (loss threshold and deadline) at each round  based on the statistical utility. For the  $R$-th round, it is defined as $U_R \leftarrow \frac{LSum_R}{L_R \cdot ddl_R}$ 
where $LSum_R$ is the loss sum of the selected samples, $L_R$ is the sum of the number of selected samples, and $ddl_R$ is the chosen deadline for the round. Note that $LSum_R$ and $L_R$ are calculated only from the clients who completed their task and succeeded in sending the model updates. The calculated $U_R$ value is added to the list $U$. \putname{} compares the $U_R$ values in the past rounds to the recent rounds (Line~\ref{algline:rc-line5}). If the past rounds have higher value than the recent rounds, \putname{} considers the model training to be stable and increases the $ltr$ value by $lss$ to further optimize the training process (Line~\ref{algline:rc-line6}). Otherwise, \putname{} decreases the $ltr$ value by $lss$ (Line~\ref{algline:rc-line9}). Note that $ddlr$, which is initialized as 1.0, is controlled in opposite direction with $dss$ (Lines~\ref{algline:rc-line7} and ~\ref{algline:rc-line10}). Such control of $ltr$ and $ddlr$ happens every $w$ round (Line~\ref{algline:rc-line4}).

\begin{algorithm}[t]
\caption{$ltr$ and $ddlr$ control at $R$-th round}\label{alg:rc}
\begin{algorithmic}[1]
\Procedure{Ctrl}{$U$, $LSum_R$, $L_R$, $ddl_R$, $ltr$, $ddlr$, $lss$, $dss$, $w$} \label{algline:rc-line1}
\State $U_R \gets \frac{LSum_R}{L_R \cdot ddl_R}$\label{algline:rc-line2}
\State $U.insert(U_R)$ \label{algline:rc-line3}
\If{$R \bmod w \equiv 0$}\label{algline:rc-line4}
\If{$\sum U(R-2w:R-w) > \sum U(R-w:R)$}\label{algline:rc-line5} 
\State $ltr \gets $\texttt{min}$(ltr + lss, 1.0)$\label{algline:rc-line6}
\State $ddlr \gets $\texttt{max}$(ddlr - dss, 0.0)$\label{algline:rc-line7}
\Else\label{algline:rc-line8}
\State $ltr \gets $\texttt{max}$(ltr - lss, 0.0)$\label{algline:rc-line9}
\State $ddlr \gets $\texttt{min}$(ddlr + dss, 1.0)$\label{algline:rc-line10}
\EndIf\label{algline:rc-line11}
\EndIf\label{algline:rc-line12}
\State \textbf{return} $ltr$, $ddlr$\label{algline:rc-line13}
\EndProcedure\label{algline:rc-line14}
\end{algorithmic}
\end{algorithm}




\subsubsection{Client selection with sample selection}
\label{sec:clientselection}

Researchers have studied on \textit{how to select a group of clients for a training round} to optimize convergence speed and model performance in heterogeneous FL~\cite{lai01, cho01, cho02}. While these approaches prioritize clients with higher statistical utility from the data, applying them along with \putname{} is non-trivial as the samples are dynamically selected with the loss threshold. To address this issue, we propose a new formulation to calculate the statistical utility of a client $i$ along with the sample selection strategy of \putname{} as follows:
\[ StatUtil(i) \leftarrow \lvert len(OT^i) \rvert \sqrt{\frac{1}{\lvert len(OT^i) \rvert}\sum_{s \in OT^i}Loss(s)^2} .\]
This is based on the formulation of statistical utility of state-of-the-art client selection method~\cite{lai01}, which we only calculate the statistical utility from the $OT^i$ group. Thus, the sum of loss squares in $OT^i$, $\sum_{s \in OT^i}Loss(s)^2$ and the number of samples in $OT^i$, $len(OT^i)$ are also collected from clients as a differentially-private metadata.

\subsection{Adaptive Deadline Control}
\label{sec:adc}

We explain how \putname{} finds an optimal deadline for each round when the clients' training time changes with the sample selection.

\subsubsection{Efficiency of a deadline}
\label{sec:ddle}

\begin{figure}[t]
    \centering
    \begin{subfigure}[b]{0.235\textwidth}
        \centering
        \includegraphics[width=\textwidth]{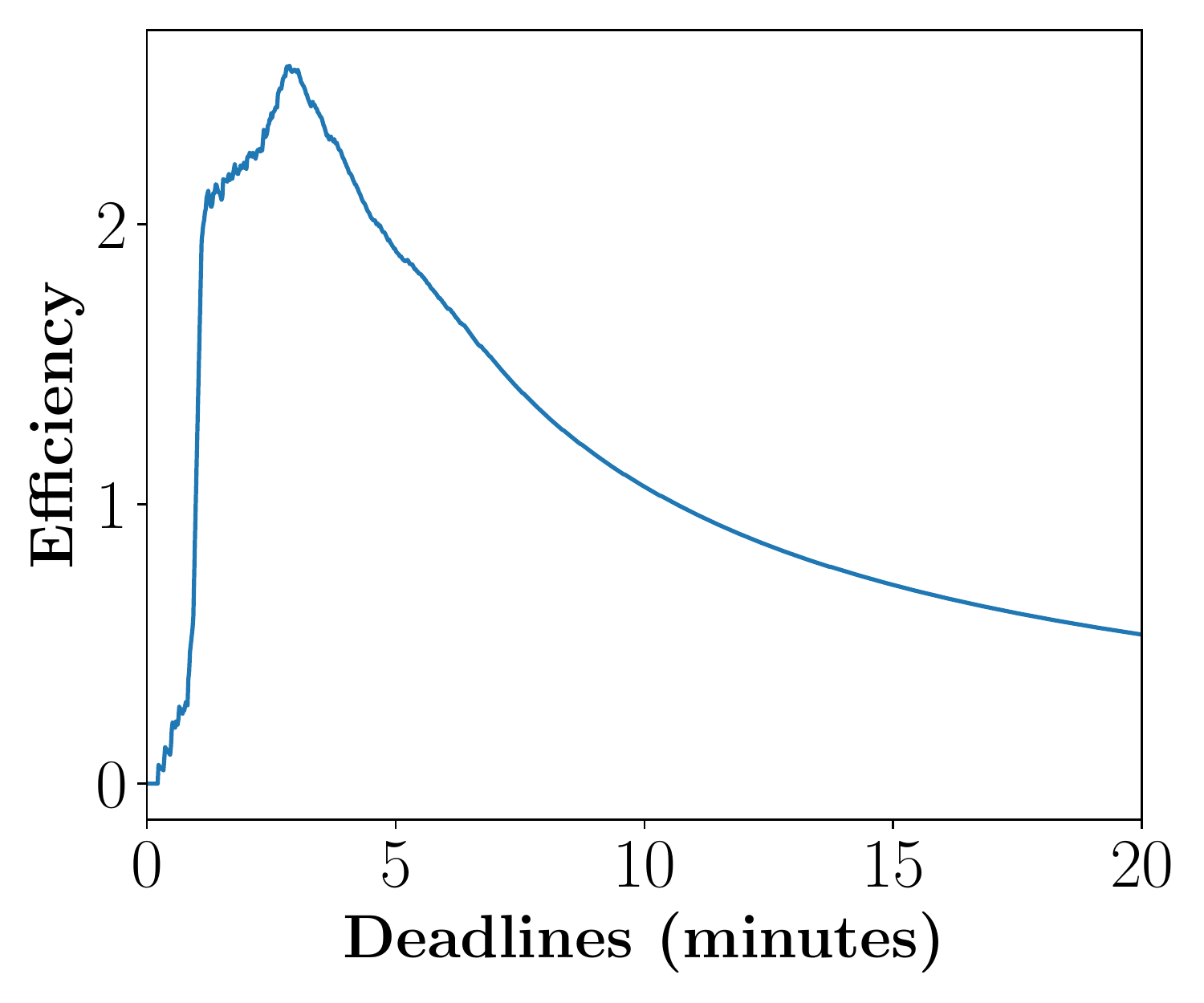}
        \caption{FEMNIST dataset.}
        \label{fig:adc-1}
    \end{subfigure}
    \hfill
    \begin{subfigure}[b]{0.235\textwidth}
        \centering
        \includegraphics[width=\textwidth]{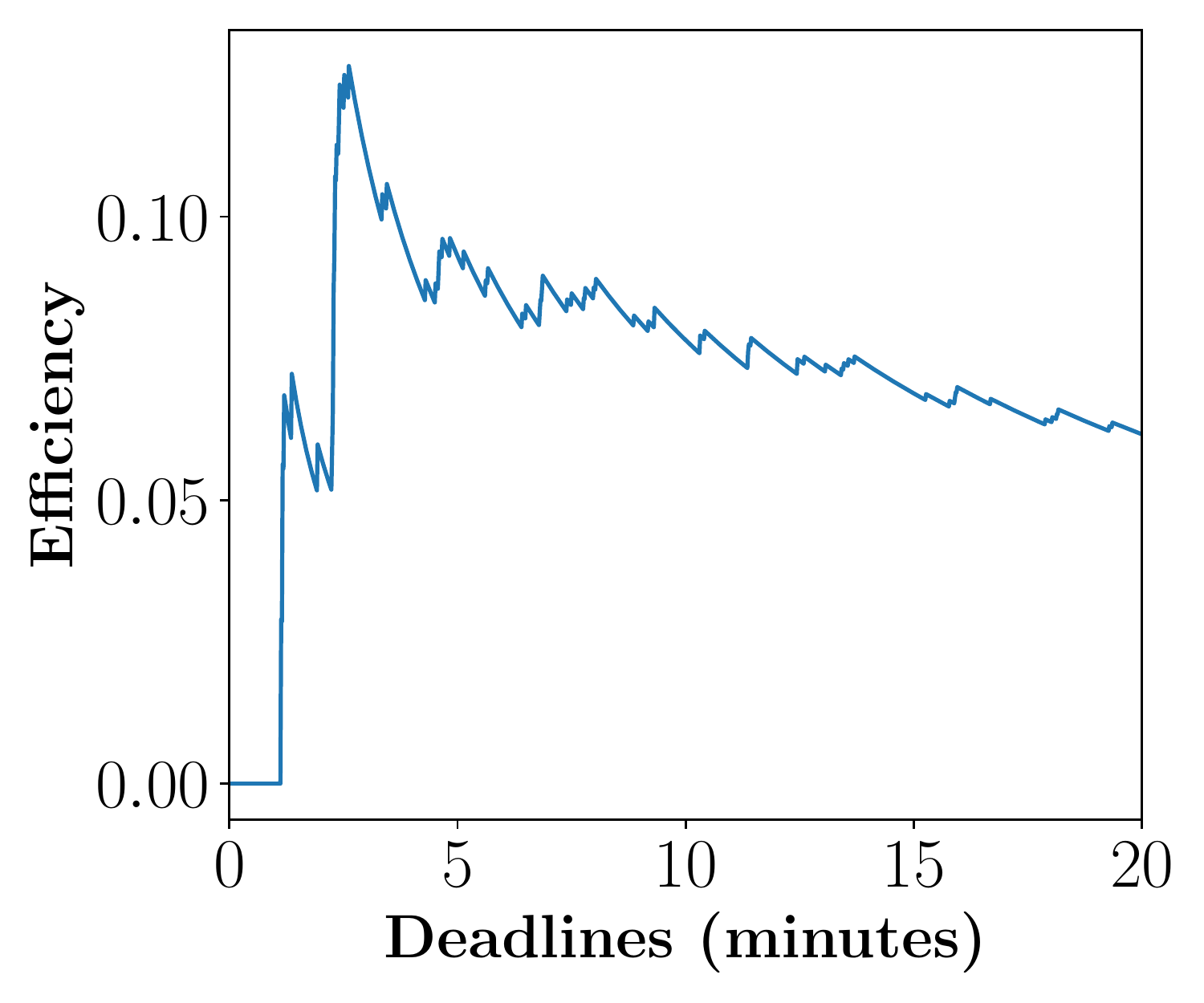}
        \caption{Shakespeare dataset.}
        \label{fig:adc-2}
    \end{subfigure}
    \vspace{-0.6cm}
  \caption{Deadline efficiency (DDL-E) evaluation on different deadlines on two FL tasks.}
  \vspace{-0.5cm}
  \label{fig:adc}
\end{figure}

In order to find the best deadline for each round, we define a metric named \textit{deadline efficiency (DDL-E)} for deadline $t$ as follows: 
\[ DDL\text{-}E(t) \leftarrow \frac{\text{\# of completed clients before } t}{t}.\]
Our definition of DDL-E formulates the benefit of using deadline $t$ by measuring the amount of completed clients per time. Finding a deadline with high DDL-E value allows the system to avoid choosing too long or too short deadlines. Setting a long deadline with a large $t$ value would have more completed clients but have low efficiency. On the other hand, configuring an extremely short deadline with a small $t$ would result in almost no completed clients and low efficiency.

To understand how DDL-E is distributed at different deadlines at FL task, we profiled the DDL-E based on a large-scale smartphone dataset
~\cite{yang01}. It contains the downlink and uplink network connectivity data and the model training latency data from real-world clients using heterogeneous hardware. We measured the DDL-E for deadlines in range of $[1sec, 1200sec]$ on two FL tasks: FEMNIST and Shakespeare. 

Figure~\ref{fig:adc} presents the DDL-E measurements on two FL tasks. From both tasks, we observe that there exists a specific deadline that shows the peak DDL-E value. From the FEMNIST dataset, $t = 172$ seconds shows the max efficiency with DDL-E value of $2.57$, while in the Shakespeare dataset, $t = 157$ seconds shows the max efficiency with DDL-E value of $0.13$. The distribution shape of DDL-E values showing sharp peak around the max value implies that finding an optimal deadline for FL with \putname{} could enable more completed clients and improved convergence speed.

\begin{algorithm}[t]
\caption{$ddl$ selection for next ($R+1$)-th round}\label{alg:ps2}
\begin{algorithmic}[1]
\Procedure{FindPeakDDL-E}{$\{DL^i, UL^i, B^i\}_{i=1}^{N}, numEpoch$}\label{algline:ps-line8}
\State $completeTime \gets \emptyset$\label{algline:ps-line9}
\State $DDL\text{-}E \gets \emptyset$ \Comment{initialize a list of DDL-E values}\label{algline:ps-line10}
\State $c \gets 0$\label{algline:ps-line11}
\State $t \gets 1$\label{algline:ps-line12}
\For{$i \gets 1$ to $N$}\label{algline:ps-line13}
\State $T_{network} \gets mean(DL^i) + mean(UL^i)$\label{algline:ps-line14}
\State $T_{train} \gets $\hspace{0.1cm}\texttt{getTrainTime}$(mean(B^i), numEpoch)$\label{algline:ps-line16}
\State $completeTime.insert(T_{network} + T_{train})$\label{algline:ps-line17}
\EndFor\label{algline:ps-line18}
\While{$c \neq N$}\label{algline:ps-line19}
\State $c \gets 0$\label{algline:ps-line20}
\For{$i \gets 1$ to $N$}\label{algline:ps-line21}
\If{$t \geq completeTime(i)$}\label{algline:ps-line22}
\State $c \gets c + 1$\label{algline:ps-line23}
\EndIf\label{algline:ps-line24}
\EndFor\label{algline:ps-line25}
\State $DDL\text{-}E.insert(\frac{c}{t})$\label{algline:ps-line26}
\State $t \gets t + 1$\label{algline:ps-line27}
\EndWhile\label{algline:ps-line28}
\State \textbf{return} \texttt{maxIndex}$(DDL\text{-}E)$\label{algline:ps-line29}
\EndProcedure\label{algline:ps-line30}
\Procedure{SelectDeadline}{$\{DL^i, UL^i, B^i\}_{i=0}^{N}, E, ddlr$}\label{algline:ps-line32}
\State $dl \gets $\textsc{FindPeakDDL-E}$(\{DL^i, UL^i, B^i\}_{i=1}^{N}, 1)$\label{algline:ps-line33}
\State $dh \gets $\textsc{FindPeakDDL-E}$(\{DL^i, UL^i, B^i\}_{i=1}^{N}, E)$\label{algline:ps-line34}
\State $ddl_{R+1} \gets dl + (dh - dl) \cdot ddlr$\label{algline:ps-line35}
\State \textbf{return} $ddl_{R+1}$\label{algline:ps-line36}
\EndProcedure\label{algline:ps-line37}
\end{algorithmic}
\end{algorithm}


We designed \putname{} to select a deadline based on finding the best DDL-E values. Lines~\ref{algline:ps-line8}- ~\ref{algline:ps-line30} of Algorithm~\ref{alg:ps2} shows how \putname{} finds the deadline with max DDL-E value. This is based on the clients' hardware capability information: for client $i$, they are DownLink speed ($DL^i$), UpLink speed ($UL^i$), and Batch training latency ($B^i$) (Line ~\ref{algline:ps-line8}). When there are a total of $N$ clients involved, we assume clients have already profiled $\{DL^i, UL^i\}_{i=1}^{N}$ before FL and \putname{} collects $\{B^i, I^i\}_{i=1}^{N}$ during FL. \putname{} first iterates over N clients to measure their completion time with the mean value of $\{DL^i, UL^i, B^i\}_{i=1}^{N}$ (Lines~\ref{algline:ps-line13}- ~\ref{algline:ps-line18}). \putname{} then measures the DDL-E from the smallest deadline $t$ (we use $t=1sec$)
and increments $t$ until all clients complete before $t$ (Lines~\ref{algline:ps-line19}-~\ref{algline:ps-line28}) and outputs the deadline with the max DDL-E value (Line~\ref{algline:ps-line29}). 

As different subset of clients are selected for each round, \putname{} finds the max DDL-E value among the selected clients of each round. Moreover, as clients use different size of training data with sample selection, we estimate the training time of client $i$ (\texttt{getTrainTime} from Line~\ref{algline:ps-line16}) as follows:
\[ \frac{len(OT^i) - 1 }{batchSize} \cdot mean(B^i) \cdot numEpoch. \]
We use the length of $OT^i$ in measuring the training time of a client to reflect the number of samples being selected. This allows \putname{} to reliably find the deadline with the best DDL-E along with the sample selection strategy.

\subsubsection{Deadline selection module}

The \textit{deadline selection module} of \putname{} determines the deadline that optimizes the training process and convergence speed. The module selects the deadline as shown in Lines~\ref{algline:ps-line32} -~\ref{algline:ps-line37} of Algorithm~\ref{alg:ps}. The module measures \textit{deadline low} value $dl$ and \textit{deadline high} value $dh$, which are the deadlines with the max DDL-E value when clients are running 1 epoch and $E$ epochs, respectively (Lines~\ref{algline:ps-line33} and ~\ref{algline:ps-line34}). 

The module then outputs the deadline of the next round (R+1) with parameter \textit{deadline ratio} ($ddlr$) that calculates the linear interpolation between $dl$ and $dh$ (Line~\ref{algline:ps-line35}) as $ddl_{R+1} \leftarrow dl + (dh - dl) \cdot ddlr$. 
The reason of selecting a value between $dl$ and $dh$ is because \putname{} is built on top of Prox~\cite{li01} that allows clients to train various number of epochs within the deadline. With an aim to optimize the training efficiency, \putname{} initially configures $ddlr$ as 1.0 and gradually decreases the value by the parameter $dss$, as explained in Section~\ref{sec:ltsm}.

\subsection{Collaboration with FL Methods}

One advantage of \putname{} is its applicability to orthogonal FL approaches that do not perform sample selection on clients or control the deadline. Applying \putname{} could be achieved by simply adding the sample selection and deadline control strategies on top of other methods. We demonstrate the collaboration capability of \putname{} with its implementation on top of three existing FL methods and improved performances in Section~\ref{sec:evaluationcollaboration}.

Some recent FL approaches, such as Oort~\cite{lai01}, use one batch instead of the full dataset for training in each local epoch, making them non-trivial to be directly integrated with \putname{}. To address this issue,
we propose \textit{OortBalancer}, which 
is built on top of \textit{Oort}, where the sample selection strategy of \putname{} is adopted with one adjustment: $L$ from Algorithm~\ref{alg:ssm} is fixed to the batch size. Intuitively, while \textit{Oort} trains one randomly selected batch for each local epoch, \textit{OortBalancer} selects samples for one batch that focuses on more important samples and thereby optimizes the training process. We demonstrate the performance of \textit{OortBalancer} in Section~\ref{sec:evaluationcollaboration} as one of the three examples of \putname{} collaboration.


\section{Evaluation}
\label{sec:evaluation}

We evaluate \putname{} to answer the following key questions: 1) How much performance improvement (in terms of time-to-accuracy and model accuracy) does \putname{} achieves over existing FL methods? 2) How sensitive is \putname{} with different choice of parameters?
3) How much performance improvement does each component of \putname{} achieves? 4) How does \putname{} perform when it jointly operates with orthogonal FL approaches?

\subsection{Experimental Setup}
\label{sec:experimentalsetup}

\begin{table*}[t]
\caption{\rev{Speedup and accuracy on five datasets with the real-world user data.}}
\label{table:speedup-and-accuracy}
\begin{center}
\scalebox{0.82}{
\begin{tabular}{ccccccccccc}
\toprule
\multicolumn{1}{c}{Task}&\multicolumn{4}{c}{CV}&\multicolumn{4}{c}{NLP}&\multicolumn{2}{c}{HAR} \\
\cmidrule(lr){2-5}
\cmidrule(lr){6-9}
\cmidrule(lr){10-11}
\multicolumn{1}{c}{Dataset} &\multicolumn{2}{c}{FEMNIST}&\multicolumn{2}{c}{Celeba}&\multicolumn{2}{c}{Reddit}&\multicolumn{2}{c}{Shakespeare}&\multicolumn{2}{c}{UCI-HAR} \\ \midrule
Methods&Speedup&Acc.&Speedup&Acc.&Speedup&Acc.&Speedup&Acc.&Speedup&Acc. \\ 
\midrule
FedAvg+1$T$&1.00$\pm$0.00&.796$\pm$.007&0.97$\pm$0.05&.851$\pm$.005&0.08$\pm$0.12&.090$\pm$.001&0.24$\pm$0.22&.399$\pm$.020&0.49$\pm$0.36&.814$\pm$.029 \\
FedAvg+2$T$&0.59$\pm$0.01&.763$\pm$.009&0.59$\pm$0.04&.824$\pm$.010&0.58$\pm$0.07&.104$\pm$.000&0.39$\pm$0.10&.373$\pm$.046&0.68$\pm$0.07&.819$\pm$.014 \\
FedAvg+SPC&0.71$\pm$0.02&.777$\pm$.004&0.80$\pm$0.14&.829$\pm$.013&0.87$\pm$0.08&.112$\pm$.001&0.56$\pm$0.10&.416$\pm$.017&0.91$\pm$0.13&.840$\pm$.016 \\
FedAvg+WFA&0.33$\pm$0.20&.594$\pm$.205&0.54$\pm$0.04&.813$\pm$.020&1.00$\pm$0.00&.113$\pm$.001&1.00$\pm$0.00&.439$\pm$.017&0.68$\pm$0.07&.819$\pm$.014 \\
Prox+1$T$&0.99$\pm$0.02&.795$\pm$.008&1.05$\pm$0.03&.855$\pm$.002&2.87$\pm$0.43&.121$\pm$.005&1.14$\pm$0.16&.476$\pm$.003&0.96$\pm$0.09&.849$\pm$.008 \\
Prox+2$T$&0.65$\pm$0.02&.767$\pm$.006&0.75$\pm$0.04&.833$\pm$.010&3.63$\pm$0.43&.127$\pm$.005&1.00$\pm$0.14&.457$\pm$.003&0.68$\pm$0.07&.819$\pm$.014 \\
SampleSelection&1.01$\pm$0.02&.799$\pm$.006&1.03$\pm$0.06&.852$\pm$.002&1.52$\pm$0.38&.118$\pm$.001&0.85$\pm$0.24&.439$\pm$.021&0.90$\pm$0.05&.845$\pm$.008 \\
\midrule
\putname{}&1.57$\pm$0.03&.815$\pm$.006&1.43$\pm$0.07&.862$\pm$.006&4.48$\pm$0.23&.146$\pm$.001&1.20$\pm$0.10&.489$\pm$.004&1.56$\pm$0.28&.855$\pm$.034\\
\putname{}-A&1.60$\pm$0.06&\textbf{.820$\pm$.003}&1.52$\pm$0.07&\textbf{.873$\pm$.004}&4.08$\pm$0.76&\textbf{.154$\pm$.000}&1.31$\pm$0.28&\textbf{.505$\pm$.004}&1.39$\pm$0.43&\textbf{.893$\pm$.008}\\
\putname{}-S&\textbf{1.71$\pm$0.01}&.819$\pm$.002&\textbf{1.67$\pm$0.04}&.859$\pm$.006&\textbf{4.99$\pm$0.42}&.148$\pm$.003&\textbf{1.83$\pm$0.14}&.488$\pm$.001&\textbf{1.98$\pm$0.48}&.863$\pm$.010\\
\bottomrule
\end{tabular}
}
\end{center}
\end{table*}


\textbf{Implementation.} We developed \putname{} on FLASH~\citep{yang01}, a heterogeneity-aware benchmarking framework for FL based on LEAF~\citep{caldas01}. FLASH provides a simulation of heterogeneous computational capabilities and network connectivity from a large-scale real-world trace dataset collected over 136k smartphones that span one thousand types of devices. We implement \putname{} with the state-of-the-art FL aggregation method Prox~\cite{li01}. Our implementation is based on Python 3.6 and TensorFlow 1.14 with 2,062 lines of code on top of FLASH. 
\rev{The source code of our \putname{} implementation are available at \url{https://github.com/jaemin-shin/FedBalancer}.}

\textbf{Datasets.} To simulate FL tasks in our evaluation, we use five datasets that contain data generated by real-world users, which are categorized in three different domains as follows:

\begin{itemize}[topsep=3pt, align=left, labelwidth=10pt, leftmargin=15pt]
\item {\textit{Computer Vision (CV)}}: For CV, we evaluated \putname{} on two image recognition datasets: FEMNIST~\cite{cohen01} and Celeba~\cite{liu01}. FEMNIST dataset contains images of handwritten digits and characters from 712 users with total 157,132 samples. Celeba dataset contains face attributes of 915 users with 19,923 samples. We use CNN models for both datasets as in previous work~\cite{yang01}.
\item {\textit{Natural Language Processing (NLP)}}: We evaluate \putname{} on two NLP tasks each on different dataset: next-word prediction on Reddit~\cite{caldas01} dataset and next-character prediction on Shakespeare~\cite{shakespeare01} dataset. The Reddit dataset contains reddit posts from 813 users with 32,680 samples, and the Shakespeare dataset contains 845,231 samples separated into 171 users. We use LSTM models for both datasets as in previous work~\cite{yang01}.
\item {\textit{Human Activity Recognition (HAR)}}: We use UCI-HAR dataset~\cite{anguita01} that contains six types of activity data on accelerometer and gyroscope from 30 users with 10,299 samples. As in previous work, we use \rev{CNN models~\cite{ek01}}.
\end{itemize}

\textbf{Metrics.} As in the previous work~\citep{lai01} that evaluated on heterogeneous FL clients, we mainly evaluate \textit{time-to-accuracy} performance and \textit{final model accuracy} on the experiments. Here, the \textit{time-to-accuracy} performance indicates the wall clock time that is required for a model training task to reach an accuracy target. We repeat each experiment for three times with different random seeds and report the average and standard deviation of these evaluation metrics.





\textbf{Baselines.} We use the following list of approaches as a baseline to compare with \putname{} in our evaluation:
\begin{itemize}[topsep=3pt, align=left, labelwidth=10pt, leftmargin=15pt]
\item {\textit{Aggregation algorithms}}: We use FedAvg~\cite{mcmahan01} and Prox~\cite{li01}, the most widely used aggregation algorithms for FL. Prox offers an optimizer with convergence guarantee on heterogeneous clients, while allowing clients to train various number of local epochs within the deadline. For the $\mu$ parameter of Prox, we tested $\mu$ in $\{0.0, 0.001, 0.01,$ $0.1, 1.0\}$ as suggested by the paper and pick one with the best final accuracy. All the datasets showed the best accuracy with $0.0$ except Celeba with $0.1$.
\item {\textit{Deadline configuration methods}}: We use four different deadline configuration methods in the evaluation. We configure two different fixed deadlines as a baseline, which are $1T$ and $2T$. Before the training begins, we sample the round completion time of all participating clients and calculate the mean value as $T$. Thus, $2T$ uses double of that mean value as a fixed deadline. We also adopt SmartPC (SPC)~\cite{li04}, and implemented it to involve the certain portion $U_{required}$ of users to complete at a training round. As suggested in the paper, we use 80\% for $U_{required}$. Lastly, we use a method that waits every client to finish a round, which we named as WaitForAll (WFA).
\item \rev{{\textit{Sample selection method}}: We implement the baseline sample selection method that is a combination of the following: (1) We determined \textit{how many} samples to select based on FedSS~\cite{cai01}, which controls the training dataset size on clients with larger datasets for each training round. (2) There were several approaches that propose \textit{which} samples to select based on \textit{loss}~\cite{loshchilov01, schaul01}, \textit{gradient}~\cite{alain01}, or \textit{gradient norm upper bound}~\cite{katharopoulos01, li08} of samples. As in \putname{}, we use \textit{loss} to select samples for the baseline experiment.}
\end{itemize}
For baseline experiments, we use the combination of the aggregation algorithms and the deadline configuration methods of above. 
We do not test SPC with Prox as it is nontrivial to accept stragglers' model update with less number of epochs when $U_{required}$ users complete a training round. Moreover, we do not test WFA with Prox as it is identical with FedAvg when $\mu = 0$, which is used by most datasets. \rev{We test the sample selection method with the best performing deadline configuration methods with Prox, which is Prox+$2T$ for Reddit dataset and Prof+$1T$ otherwise.}

\begin{table*}[t]
\caption{\rev{$\{w, lss, dss, p\}$ parameters from the \putname{} experiments in Table~\ref{table:speedup-and-accuracy}. Parameters with forward slashes indicate the different parameters from the multiple runs of experiments with three different random seeds.}}
\vspace{-0.2cm}
\label{table:parameters}
\begin{center}
\scalebox{0.815}{
\begin{tabular}{c|cccc|cccc|cccc}
\toprule
\multicolumn{1}{c}{}&\multicolumn{4}{c}{\putname{}}&\multicolumn{4}{c}{\putname{}-A}&\multicolumn{4}{c}{\putname{}-S}\\
\cmidrule(lr){2-5}
\cmidrule(lr){6-9}
\cmidrule(lr){10-13}
\multicolumn{1}{c}{}&\multicolumn{1}{c}{$w$}&\multicolumn{1}{c}{$lss$}&\multicolumn{1}{c}{$dss$}&\multicolumn{1}{c}{$p$}&\multicolumn{1}{c}{$w$}&\multicolumn{1}{c}{$lss$}&\multicolumn{1}{c}{$dss$}&\multicolumn{1}{c}{$p$}&\multicolumn{1}{c}{$w$}&\multicolumn{1}{c}{$lss$}&\multicolumn{1}{c}{$dss$}&\multicolumn{1}{c}{$p$}\\
\midrule
FEMNIST&&&&&20&0.01/0.1/0.1&0.10&1.00&20&0.05/0.05/0.01&0.25/0.1/0.05& 1.00\\
Celeba&&&&&20/5/5&0.05/0.1/0.01&0.1/0.1/0.05&1.00/0.75/1.00&20/20/5&0.01/0.01/0.1&0.1/0.1/0.25&1.00/1.00/0.75\\
Reddit&20&0.05&0.05&1.00&5/20/20&0.01&0.1/0.25/0.1&1.00&5&0.1& 0.25/0.1/0.05&0.75\\
Shakespeare&&&&&5&0.01/0.05/0.01&0.1/0.25/0.25&1/0.75/1&5/20/5&0.05/0.01/0.1&0.1/0.05/0.1&0.75\\
UCI-HAR&&&&&5/5/20&0.1/0.01/0.05&0.1/0.1/0.25&1/1/0.75&5&0.1/0.1/0.05&0.25&0.75/0.75/1\\
\bottomrule
\end{tabular}
}
\end{center}
\vspace{-0.3cm}
\end{table*}

\begin{table*}[t]
\vspace{0.2cm}
\caption{\rev{Speedup and accuracy on different choice of \putname{} parameters.}}
\vspace{-0.2cm}
\label{table:paramter-analysis}
\begin{center}
\scalebox{0.83}{
\begin{tabular}{cccc|ccc|ccc|cc}
\toprule
\multicolumn{2}{c}{\multirow{2}{*}{Parameter}}&\multicolumn{2}{c}{$w$}&\multicolumn{3}{c}{$lss$}&\multicolumn{3}{c}{$dss$}&\multicolumn{2}{c}{$p$} \\
\cmidrule(lr){3-4}
\cmidrule(lr){5-7}
\cmidrule(lr){8-10}
\cmidrule(lr){11-12}
&&$5$&$20$&$0.01$&$0.05$&$0.10$&$0.05$&$0.10$&$0.25$&$0.75$&$1.00$ \\ \midrule
\multirow{2}{*}{FEMNIST}&Speedup&1.39$\pm$0.16&\textbf{1.52$\pm$0.15}&1.37$\pm$0.15&1.49$\pm$0.18&\textbf{1.50$\pm$0.14}&\textbf{1.54$\pm$0.07}&1.47$\pm$0.16&1.35$\pm$0.18&1.41$\pm$0.15&\textbf{1.50$\pm$0.17} \\
&Accuracy&.813$\pm$.005&\textbf{.816$\pm$.004}&.813$\pm$.004&.815$\pm$.006&\textbf{.816$\pm$.006}&\textbf{.817$\pm$.003}&.814$\pm$.004&.812$\pm$.005&.814$\pm$.004&\textbf{.815$\pm$.005} \\
\midrule
\multirow{2}{*}{Reddit}&Speedup&3.40$\pm$0.53&\textbf{3.70$\pm$0.23}&3.60$\pm$0.29&\textbf{3.61$\pm$0.31}&3.44$\pm$0.61&\textbf{3.66$\pm$0.44}&3.60$\pm$0.33&3.40$\pm$0.48&\textbf{3.71$\pm$0.30}&3.39$\pm$0.49 \\ 
&Accuracy&.146$\pm$.007&\textbf{.147$\pm$.003}&\textbf{.150$\pm$.002}&.147$\pm$.002&.143$\pm$.008&\textbf{.147$\pm$.004}&.146$\pm$.007&.147$\pm$.005&\textbf{.149$\pm$.002}&.144$\pm$.007 \\
\bottomrule
\end{tabular}
}
\end{center}
\end{table*}



\textbf{Method.} We first ran FedAvg+$1T$ on five datasets until convergence, with the number of rounds that are suggested by previous works~\cite{caldas01, yang01, ek01, sozinov01}: 1000, 100, 600, 40, and 50 rounds for FEMNIST, Celeba, Reddit, Shakespeare, and UCI-HAR. Based on the user trace data of FLASH, we measured the wall clock time which FedAvg+$1T$ ran for each dataset, and ran experiments with other baselines and \putname{} until the same wall clock time. Among the four FedAvg baselines, we pick the one with best accuracy for each dataset and configure it as the target accuracy of that task. Then, we measured the speedup of other methods in achieving the target accuracy and their final model accuracy achieved within the same wall clock time.

We tested the following combination of parameters for \putname{}: $w$ in $\{5,20\}$, $lss$ in $\{0.01, 0.05, 0.10\}$, $dss$ in $\{$0.05, 0.10, 0.25$\}$, and $p$ in $\{0.75, 1.00\}$. Among the results, we report the performance of \putname{} with only one set of parameter: $\{w, lss, dss, p\} = \{20, 0.05, 0.05, 1.00\}$. This is a set of parameters that we recommend FL developers to try with their task when they are not knowledgeable of which parameter performs the best. We used this parameter for all the experiments in Section~\ref{sec:evaluation}. In Section~\ref{sec:speedupandaccuracy}, we also report the parameter set with the best final accuracy for each dataset 
as \putname{}-A \rev{and the best speedup as \putname{}\textit{-S} 
to demonstrate the maximum performance \putname{} could achieve.}

\textbf{Other configurations.} As in previous study~\cite{yang01}, we use batch size of 100 for Shakespeare and 10 for rest of the datasets. We select 100 clients at each round for datasets with more than 500 users, and otherwise select 10 users for Shakespeare and 5 users for UCI-HAR. We configured the clients to train five local epochs per round. We use learning rate of 0.001 for FEMNIST and Celeba, 2 for Reddit, 0.8 for Shakespeare, and 0.005 for UCI-HAR.

\subsection{Speedup and Accuracy on Five FL Tasks}
\label{sec:speedupandaccuracy}

Table~\ref{table:speedup-and-accuracy} shows the performance of \putname{} on five datasets compared with the baseline methods, and Table~\ref{table:parameters} shows the parameters used for \putname{}, \putname{}-A, and \rev{\putname{}-S} for each dataset. We observed that \putname{} shows improved time-to-accuracy performance over the baselines on every dataset: \putname{} reaches the target accuracy \rev{1.43$\sim$1.57$\times$} faster than the FedAvg-based baselines on CV datasets, \rev{1.36$\sim$1.58$\times$} faster than the Prox-based baselines\rev{, and 1.39$\sim$1.55$\times$ faster than the \textit{SampleSelection} baseline}. \putname{} achieves \rev{1.20$\sim$4.48$\times$} speedup and \rev{1.05$\sim$1.23$\times$} speedup compared with the FedAvg and Prox-based methods respectively\rev{, while achieving 1.41$\sim$2.95$\times$ speedup over the \textit{SampleSelection} baseline} on NLP datasets. \putname{} achieves speedup of \rev{1.56$\times$, 1.63$\times$, and 1.73$\times$} compared with FedAvg, Prox, and \textit{SampleSelection} baselines on a HAR dataset, respectively.

We noticed that \putname{} consistently shows high time-to-accuracy performance on all datasets, while the performance of baselines was inconsistent across datasets. For example, among FedAvg-based baselines, FedAvg+$1T$ shows the best time-to-accuracy performance on CV tasks but shows extremely low performance on NLP tasks. In contrast, FedAvg+WFA shows the best time-to-accuracy performance on NLP tasks among FedAvg-based baselines but shows low performance on CV tasks. In UCI-HAR, FedAvg+$1T$ and FedAvg+SPC resulted in the best performance at different experiments with different random seeds. Prox+$1T$ shows the best performance among baselines on Celeba, Shakespeare, and UCI-HAR, but shows worse performance than \rev{\textit{SampleSelection}} and Prox+$2T$ on FEMNIST and Reddit respectively.

We observed that \putname{} achieves this improvement of time-to-accuracy performance without sacrificing model accuracy; in terms of the final model accuracy, \putname{} showed improvement over the baselines on all datasets. Compared to the FedAvg-based methods, \putname{} achieved \rev{1.1$\sim$5.0\%} accuracy improvement on different datasets. \putname{} achieved \rev{0.6$\sim$3.2\%} and \rev{1.0$\sim$5.0\%} accuracy improvement over Prox-based methods and the \textit{Sampleselection} baseline. 
\putname{}-A, which marks the best accuracy of \putname{}, also shows time-to-accuracy performance improvement over baselines at all datasets --- showing further improvement in speedup on FEMNIST, Celeba, and Shakespeare. \rev{On the other hand, \putname{}-S, which reports the best time-to-accuracy performance of \putname{}, shows accuracy improvement over baselines at all datasets, with further improvement in accuracy on FEMNIST, Reddit, and UCI-HAR.} This suggests that the performance of \putname{} could be further improved with a carefully selected parameter for an FL task, while the fixed parameter set we recommend still shows improved performance.

\subsection{Parameter Sensitivity Analysis}
\label{sec:parametersensitivityanalysis}


Table~\ref{table:paramter-analysis} shows the time-to-accuracy performance and final accuracy of \putname{} on different choice of parameters $\{w, lss, dss, p\}$. For each type of parameter, we fixed it to a certain value and averaged the performance from experiments with different combination of other parameters. We used speedup compared to the best FedAvg-based baseline to measure the time-to-accuracy performance. We chose FEMNIST and Reddit to explore the effect of different parameters at different domains of FL tasks (CV and NLP).



For $w$, \putname{} shows similar final accuracy (\rev{81.3\% and 81.6\%}) performance on both of the candidate values on FEMNIST, but reports better time-to-accuracy performance (\rev{1.52$\times$ over 1.39$\times$}) with $w=20$. On Reddit, $w=20$ showed better speedup (\rev{3.71$\times$ over 3.42$\times$}) and accuracy (\rev{14.8\% over 13.3\%}). In terms of $lss$, \putname{} achieves faster training and higher accuracy when $lss\geq0.05$ on FEMNIST, but achieves better performance when $lss\leq0.05$ on Reddit. With $dss$, \putname{} performs the better when $dss\leq0.1$ on both datasets, where both showed the best time-to-accuracy and accuracy with smaller  $dss=0.05$. \rev{Lastly, \putname{} performs better with $p=1.00$ on FEMNIST but reports better performance with $p=0.75$ on Reddit. For the different trend of performance with $lss$ and $p$ parameters on two datasets, we suspect that Reddit requires more rounds with data sampled from full dataset in the early stage of training to achieve better performance, and big $lss$ and $p=1.00$ performs worse as it might quickly remove the low-loss samples from training. On the other hand, FEMNIST training performs better with high-loss samples from the early stage of training.} This suggests that the best set of parameters could be selected based on how the training at an FL task proceeds.

\begin{figure}[t]
    \centering
    \begin{subfigure}[b]{0.235\textwidth}
        \centering
        \includegraphics[width=\textwidth]{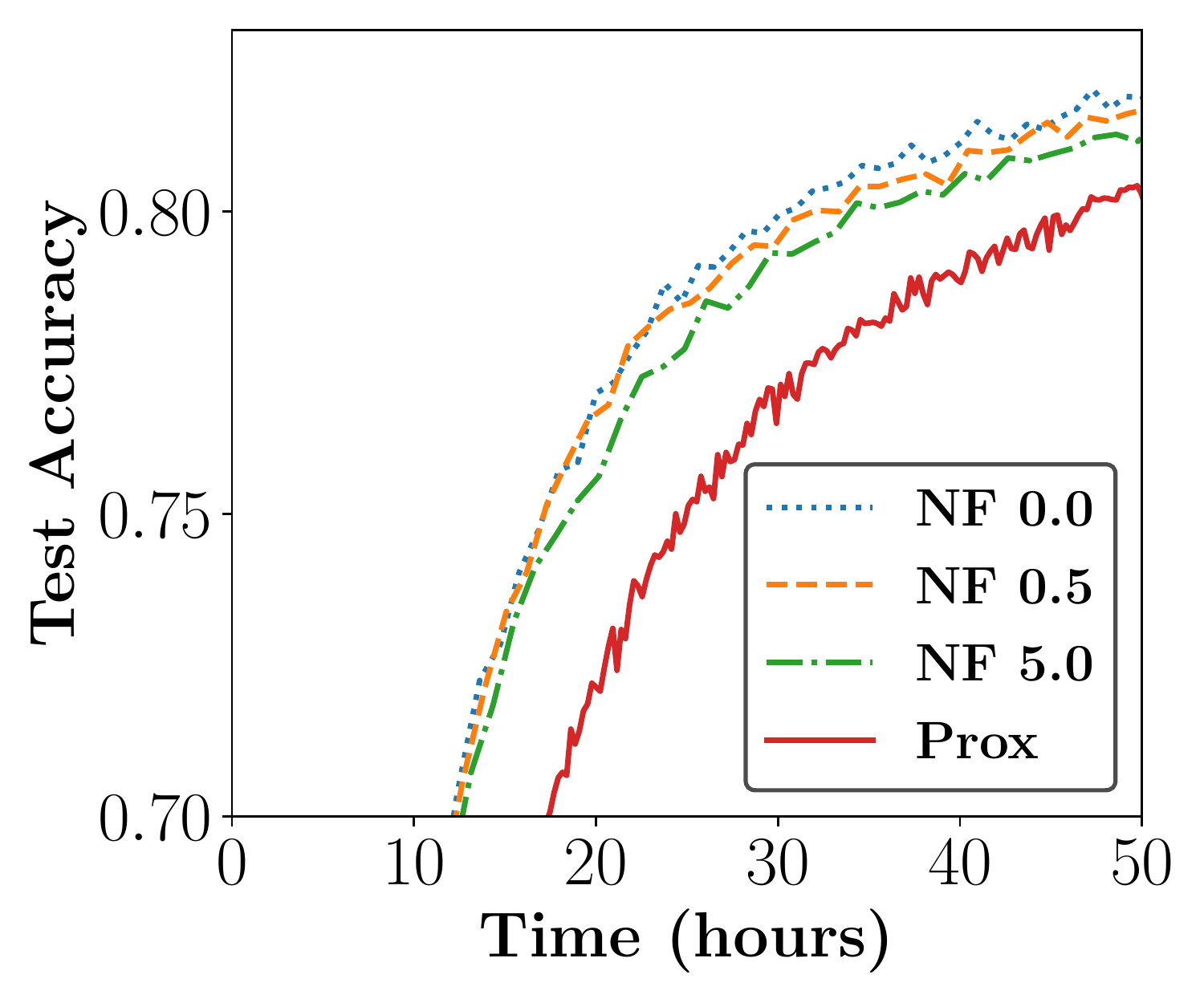}
        \caption{On Different Level of Differential Privacy.}
        \label{fig:differentialprivacy}
    \end{subfigure}
    \hfill
    \begin{subfigure}[b]{0.235\textwidth}
        \centering
        \includegraphics[width=\textwidth]{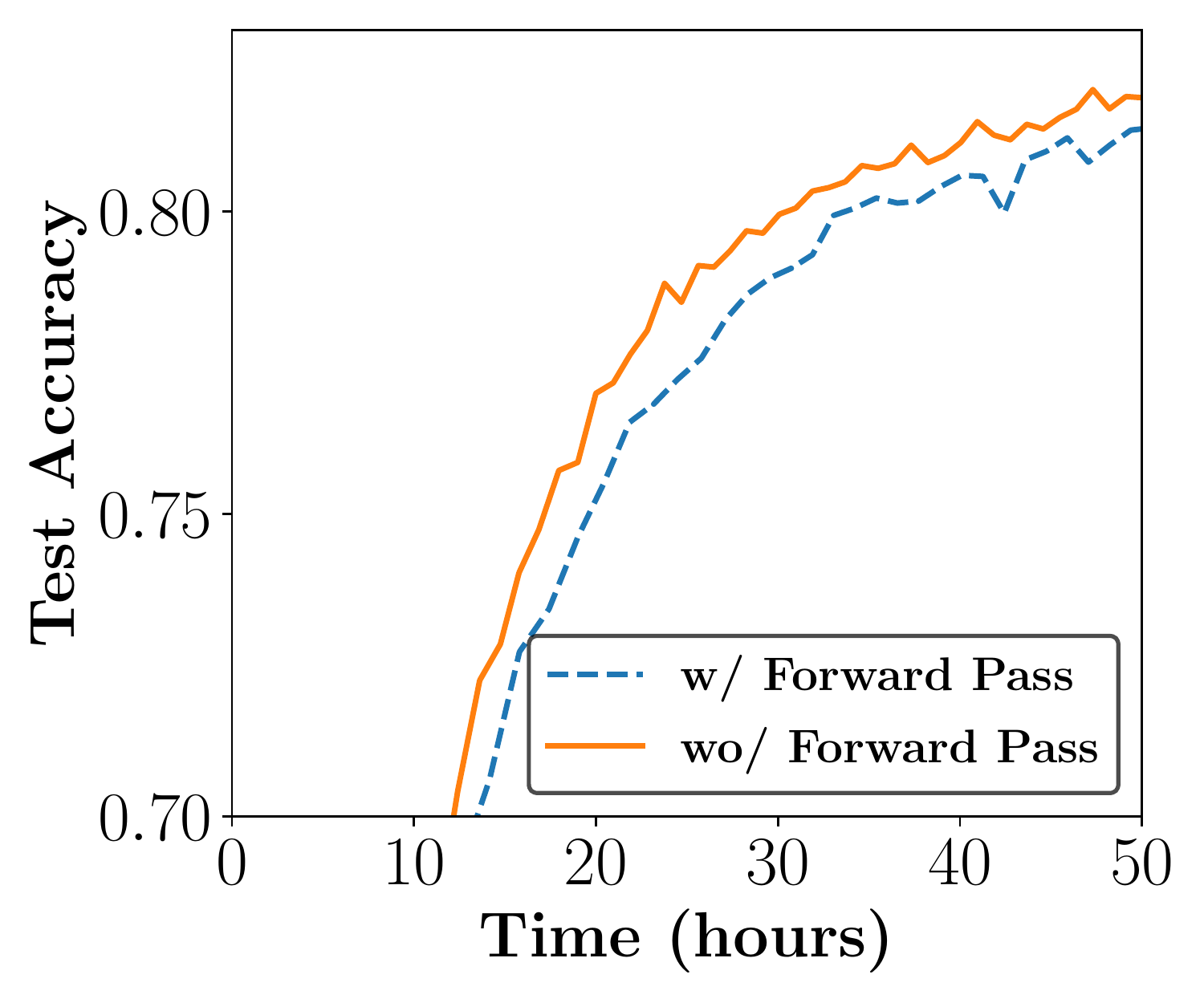}
        \caption{With and Without the Forward Pass at Each Round.}
        \label{fig:forwardpass}
    \end{subfigure}
  \caption{Performance evaluation of \putname{} with different options on FEMNIST.}
  \vspace{-0.2cm}
  \label{fig:discussion}
  \vspace{-0.2cm}
\end{figure}

Other than the algorithm parameter of \putname{}, we study the effect of different level of noise on differential privacy that we applied to mask the metadata shared by the clients. Our implementation of differential privacy is based on the previous work~\cite{lai01}, adding the noise drawn from Gaussian distribution on the metadata, with the mean as zero and the standard deviation as Noise Factor (NF). Figure~\ref{fig:differentialprivacy} shows the effect of different NFs on the performance of \putname{}. We observe that the performance of \putname{} degrades as the NF increases, as NF 0.0 achieves 81.9\% accuracy but NF 0.5 and 5.0 each achieves 81.4\% and 81.1\% accuracy. However, NF 0.5 and 5.0 still achieves better time-to-accuracy performance and accuacy compared to Prox, while NF of 5.0 is considered to be very large noise~\cite{abadi01}. This result implies that \putname{} could achieve performance improvement over the baselines while applying the differential privacy.




\subsection{Effect of \putname{} Components}
\label{sec:ablationstudy}

\begin{figure}[t]
    \centering
    \begin{subfigure}[b]{0.235\textwidth}
        \centering
        \includegraphics[width=\textwidth]{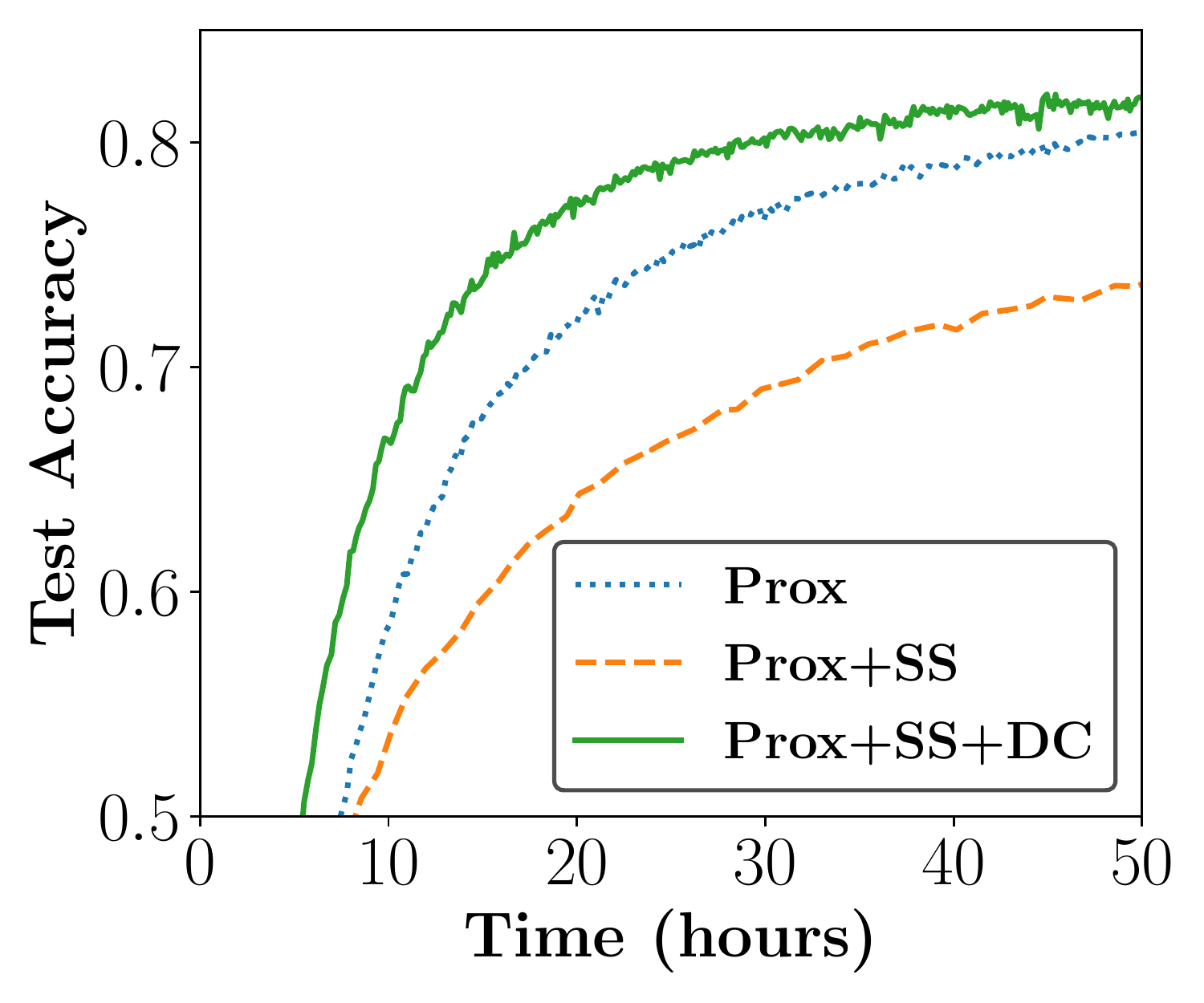}
        \caption{FEMNIST dataset.}
        \label{fig:ablation-femnist}
    \end{subfigure}
    \hfill
    \begin{subfigure}[b]{0.235\textwidth}
        \centering
        \includegraphics[width=\textwidth]{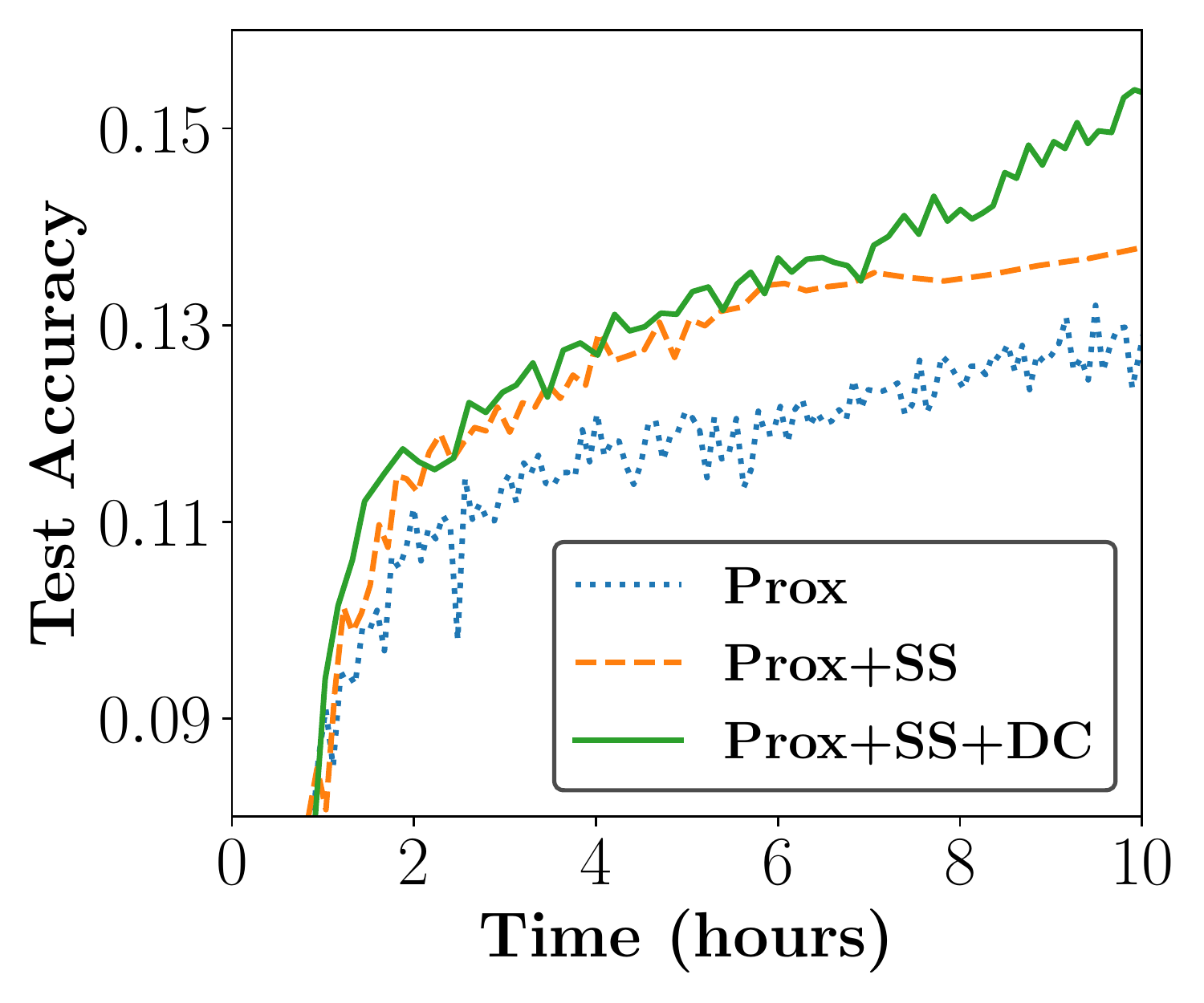}
        \caption{Reddit dataset.}
        \label{fig:ablation-reddit}
    \end{subfigure}
  \caption{Performance breakdown of \putname{} into Sample Selection (SS) and Deadline Control (DC).}
  \vspace{-0.2cm}
  \label{fig:ablation}
\end{figure}

We conducted an experiment to understand the performance brought by each component of \putname{}: Sample Selection (SS) and Deadline Control (DC). As \putname{} is built on top of Prox~\cite{li01}, we add the components one by one to observe how the performance changes when each component is introduced. 

Figure~\ref{fig:ablation} reports the result of the experiment on FEMNIST and Reddit dataset. On FEMNIST dataset, we observe that the performance drops when SS is introduced, but gets further improved when DC is added. On Reddit dataset, however, the accuracy escalates as each component is added. The reason of performance degradation on FEMNIST with SS is due to the reduced statistical utility trained at each round with the same deadline. In contrast on Reddit dataset, the performance improved with SS as it allowed more clients to successfully send their model updates within the deadline with selected client data. Moreover, we suspect that SS has brought more performance improvement on Reddit than FEMNIST by effectively selecting more important samples and the clients that have such data. The result with DC on both dataset shows its effectiveness with SS in time-to-accuracy performance improvement. 

\subsection{Collaboration with FL Algorithms}
\label{sec:evaluationcollaboration}

\begin{figure*}[t]
    \centering
    \begin{subfigure}[b]{0.32\textwidth}
        \centering
        \includegraphics[width=\textwidth]{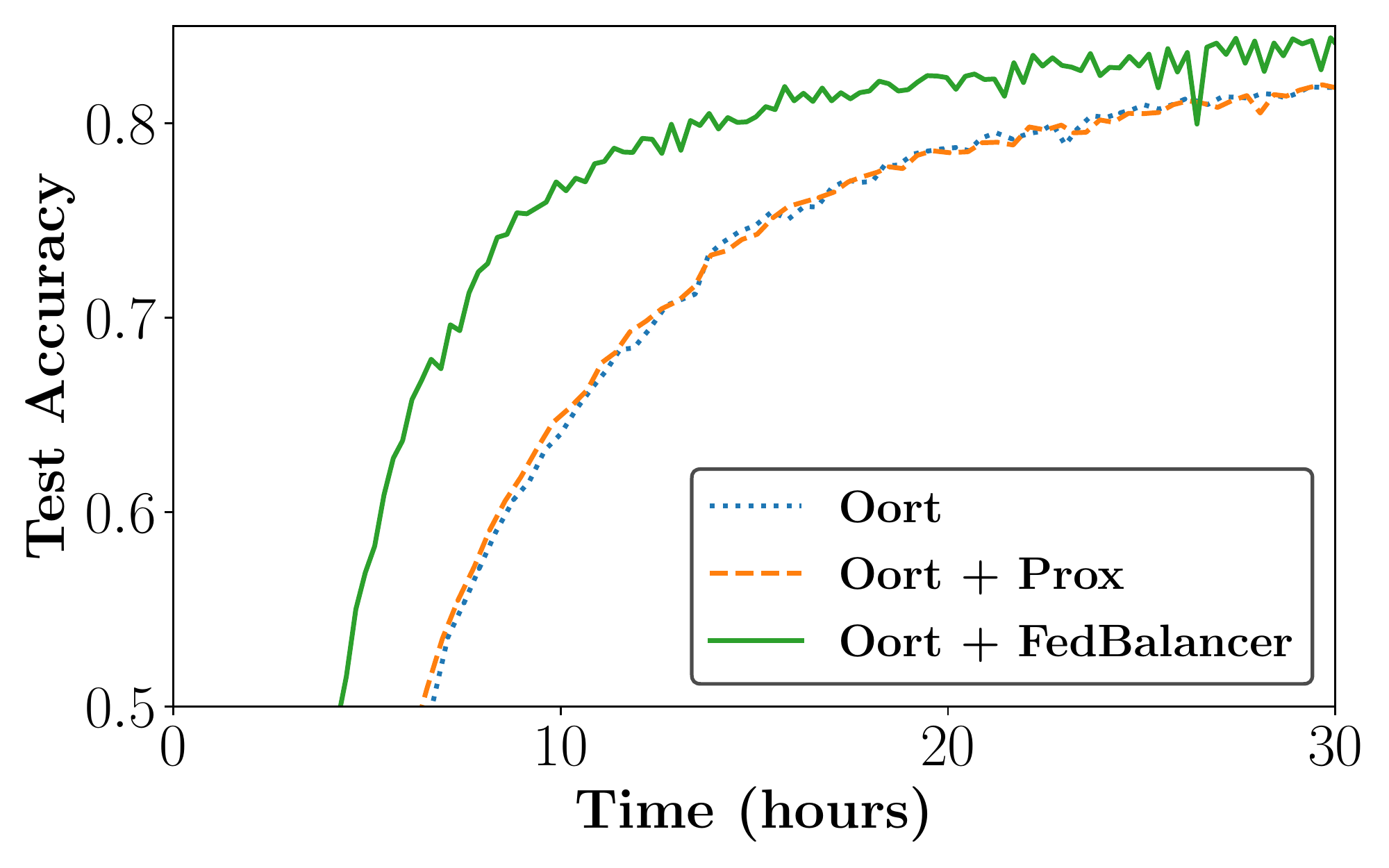}
        \caption{Oort~\cite{lai01}.}
        \label{fig:collaboration-oort}
    \end{subfigure}
    \hfill
    \begin{subfigure}[b]{0.32\textwidth}
        \centering
        \includegraphics[width=\textwidth]{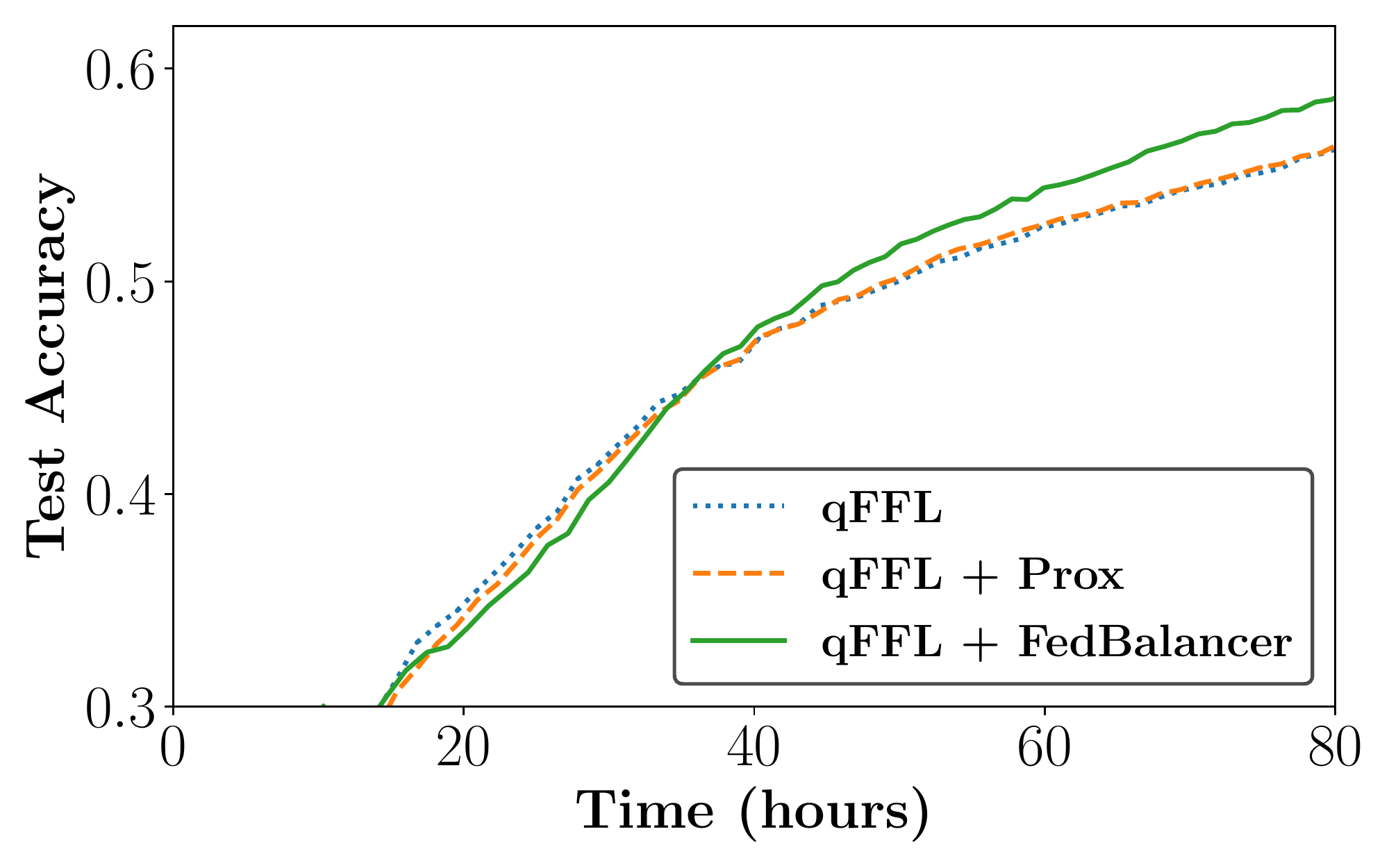}
        \caption{q-FFL~\cite{li05}.}
        \label{fig:collaboration-qffl}
    \end{subfigure}
    \hfill
    \begin{subfigure}[b]{0.32\textwidth}
        \centering
        \includegraphics[width=\textwidth]{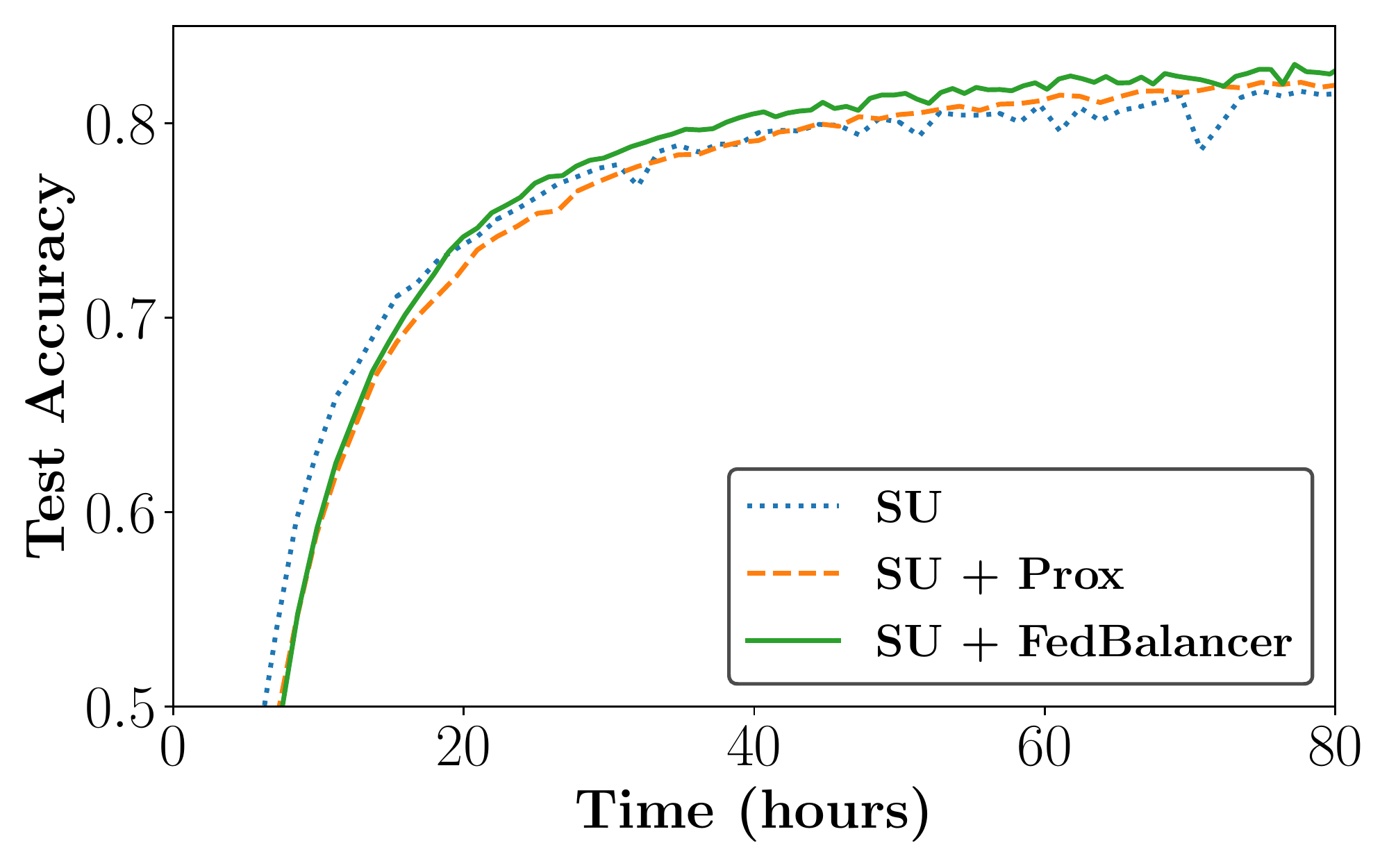}
        \caption{Structured Updates~\cite{konevcny01}.}
        \label{fig:collaboration-su}
    \end{subfigure}
    \vspace{-0.2cm}
  \caption{Collaboration of \putname{} with three FL algorithms on FEMNIST dataset.}
  \vspace{-0.2cm}
  \label{fig:collaboration}
\end{figure*}

To demonstrate the applicability of \putname{} on orthogonal FL algorithms, we implement \putname{} on top of three widely used FL approaches from different categories: \textit{Oort}~\cite{lai01} as a client selection algorithm, \textit{q-FFL}~\cite{li05} as an aggregation algorithm, and \textit{Structured Updates}~\cite{konevcny01} as a gradient compression algorithm. 
Figure~\ref{fig:collaboration} reports the experiment result, which we observe improvement in time-to-accuracy performance and model accuracy from all three cases. Collaboration with \putname{} achieved 1.84$\times$, 1.19$\times$, and 1.31$\times$ speedup, while achieving 2.6\%, 2.5\%, 1.7\% accuracy improvement on three algorithms respectively. These results suggest that \putname{} could be implemented on top of various advanced FL algorithms to achieve further performance improvement.


\begin{table}[t]
\caption{\rev{Android devices for the testbed experiments.}}
\vspace{-0.2cm}
\label{table:android}
\begin{center}
\scalebox{0.83}{
\begin{tabular}{ccccc}
\toprule
Builder&Year&Device&Processor&Quantity\\
\midrule
\multirow{5}{*}{Google}&2016&Pixel&Snapdragon 821&2\\
&2017&Pixel 2&Snapdragon 835&1\\
&2017&Pixel 2 XL&Snapdragon 835&2\\
&2018&Pixel 3&Snapdragon 845&1\\
&2020&Pixel 5&Snapdragon 765G&1\\
\midrule
\multirow{3}{*}{Samsung}&2016&Galaxy S7&Exynos 8890&1\\
&2017&Galaxy J7&Exynos 7870&1\\
&2019&Galaxy Fold&Snapdragon 855&1\\
\midrule
\multirow{2}{*}{Huawei}&2015&Nexus 6P&Snapdragon 810&3\\
&2018&P20 Lite&Kirin 659&1\\
\midrule
\multirow{1}{*}{Motorola}&2014&Nexus 6&Snapdragon 805&2\\
\midrule
\multirow{1}{*}{LG}&2015&Nexus 5X&Snapdragon 808&4\\
\midrule
\multirow{1}{*}{Essential}&2017&Essential Phone&Snapdragon 835&1\\
\bottomrule
\end{tabular}
}
\end{center}
\vspace{-0.3cm}
\end{table}

\subsection{\rev{Testbed Experiments with Android Clients}}
\label{sec:realevaluation}

\begin{table}[t]
\vspace{0.2cm}
\caption{\rev{Results from the testbed experiments.}}
\vspace{-0.2cm}
\label{table:testbedresults}
\begin{center}
\scalebox{0.85}{
\begin{tabular}{ccc}
\toprule
Method&Speedup&Accuracy\\
\midrule
FedAvg+1T&0.99$\pm$0.03&.852$\pm$.020\\
FedAvg+2T&0.75$\pm$0.30&.800$\pm$.034\\
FedAvg+SPC&0.61$\pm$0.35&.849$\pm$.013\\
FedAvg+WFA&0.92$\pm$0.07&.846$\pm$.007\\
FedProx+1T&1.03$\pm$0.23&.860$\pm$.007\\
FedProx+2T&0.90$\pm$0.20&.860$\pm$.013\\
SampleSelection&0.99$\pm$0.13&.846$\pm$.014\\
\midrule
\putname{}&\textbf{1.33$\pm$0.08}&\textbf{.885$\pm$.017}\\
\bottomrule
\end{tabular}
}
\end{center}
\vspace{-0.3cm}
\end{table}

\begin{figure}[t]
    \centering
    \begin{subfigure}[b]{0.235\textwidth}
        \centering
        \includegraphics[width=\textwidth]{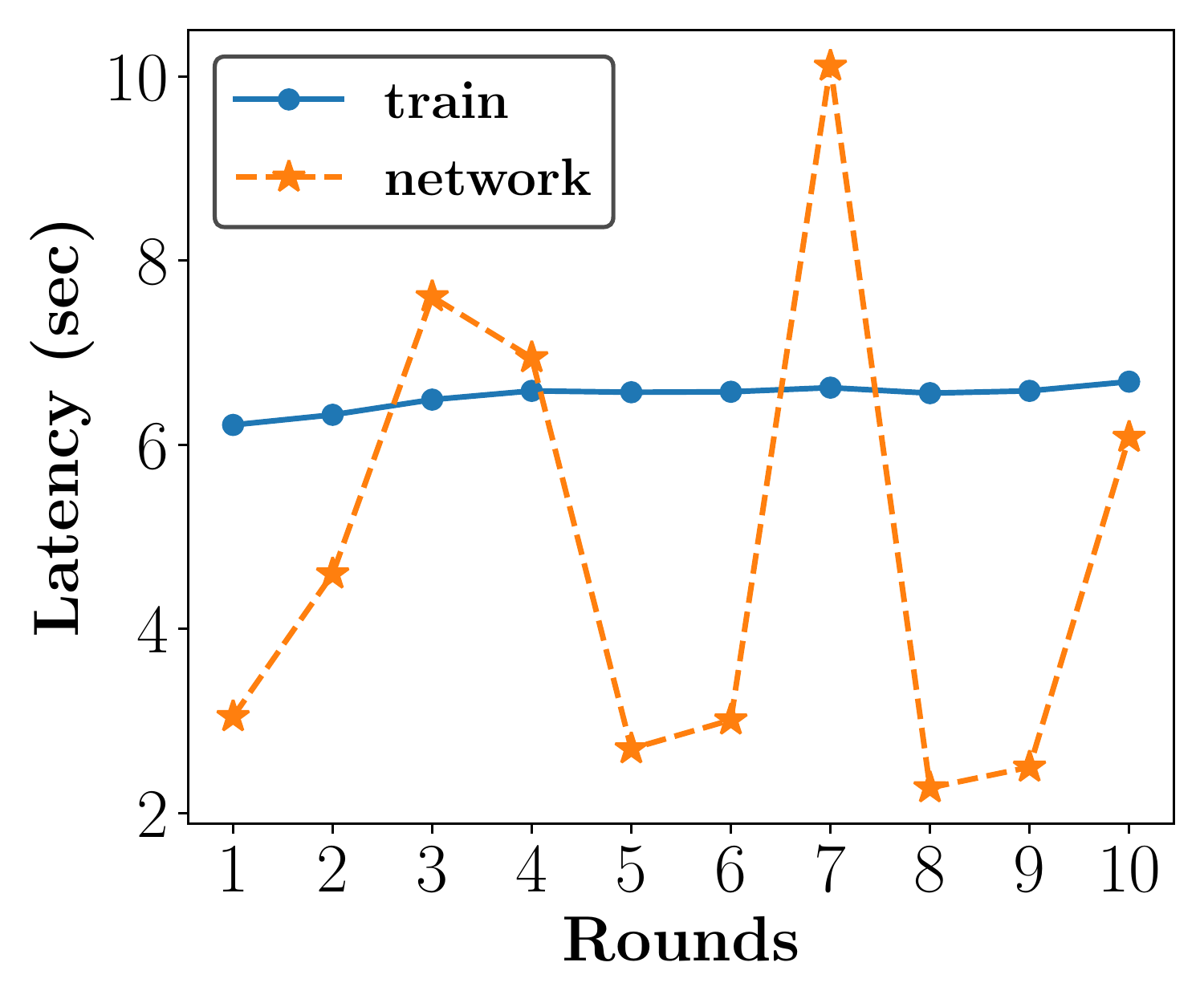}
        \caption{\rev{Google Pixel 5.}}
        \label{fig:pixel5}
    \end{subfigure}
    \hfill
    \begin{subfigure}[b]{0.235\textwidth}
        \centering
        \includegraphics[width=\textwidth]{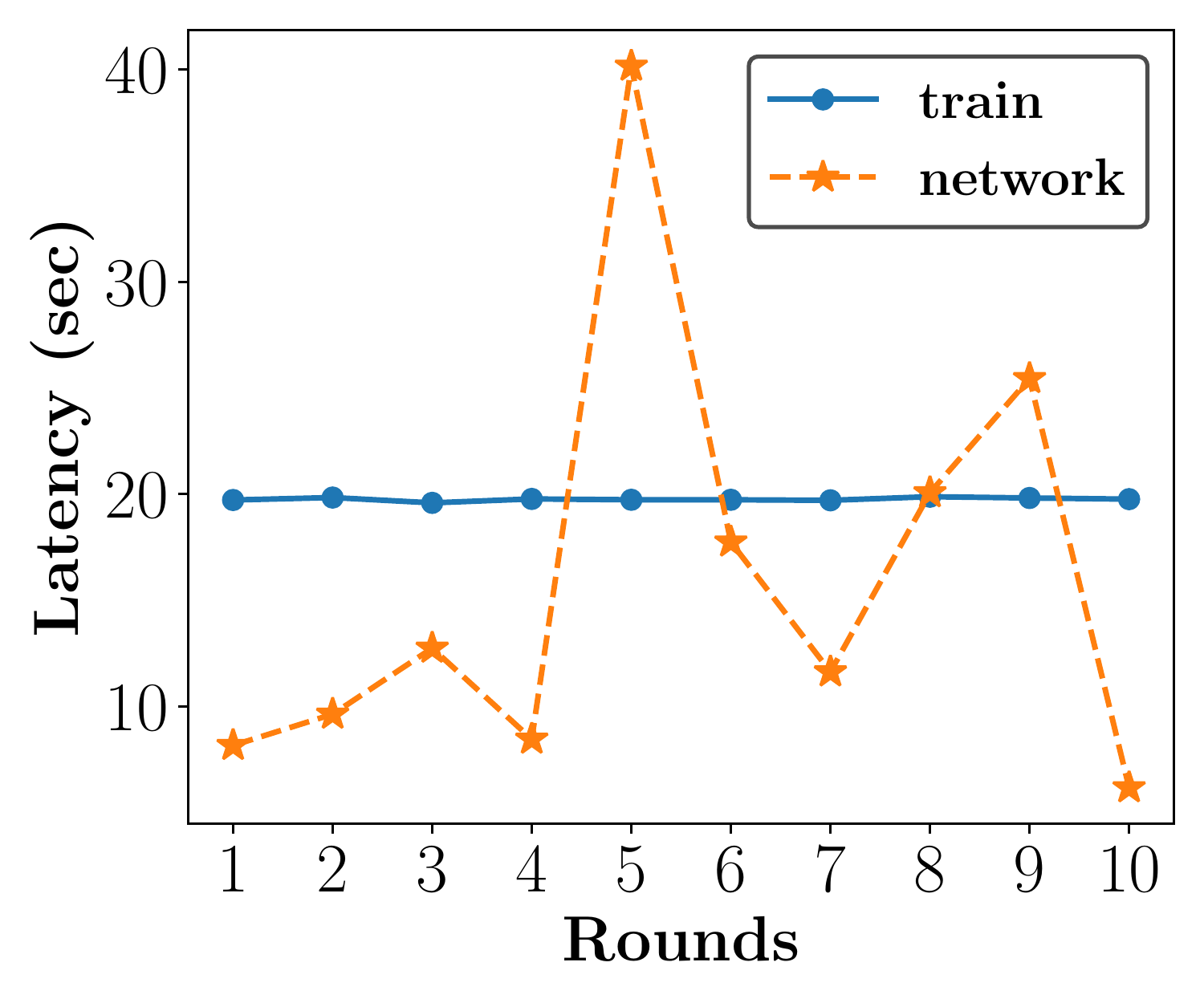}
        \caption{\rev{Huawei P20 Lite.}}
        \label{fig:p20lite}
    \end{subfigure}
\vspace{-0.4cm}
  \caption{\rev{On-device training latency (train) and communication latency (network) on different devices.}}
  \vspace{-0.2cm}
  \label{fig:devicelatency}
\end{figure}

\rev{We further conducted experiments on Android clients to understand the effectiveness of \putname{} on real hardware devices. We implemented the server using Flower~\cite{beutel01}, which is an open-source FL framework that communicates with clients via gRPC (Google Remote Procedure Call) and Protocol Buffers~\cite{protocolbuffers}. We used Ubuntu 18.04 server with Intel Xeon Gold 6254 Processor @ 3.10GHz and 512GB RAM. On-device training on Android devices were implemented with the model personalization feature of Tensorflow Lite~\cite{tflite}. We used UCI-HAR dataset and used the same experimental configuration as illustrated in Section~\ref{sec:experimentalsetup}.}

\rev{For 21 client devices in UCI-HAR experiments, we used 13 different Android models to simulate the hardware heterogeneity in the real-world as illustrated in Table~\ref{table:android}. We placed the devices in an office room of a laboratory building, where the devices were connected to a campus Wi-Fi. To configure fixed deadlines for the deadline configuration methods $1T$ and $2T$, we sampled the training round completion time of each device 10 times before the experiments.}

\rev{Table~\ref{table:testbedresults} shows the performance of \putname{} compared with the baseline methods in our testbed experiments.  \putname{} shows higher time-to-accuracy performance and final model accuracy over the baselines. \putname{} showed 1.34$\times$ speedup and 3.3\% accuracy improvement over the FedAvg-based baselines. \putname{} also achieved the target accuracy 1.29$\times$ and 1.34$\times$ faster than Prox and \textit{SampleSelection} baselines, with 2.5\% and 3.9\% accuracy improvements.}





\rev{One of the unique challenges we discovered in the testbed experiments was dynamically changing round completion times of the client devices. To further understand the cause of such a phenomenon, we separately sampled the on-device training latency and the communication latency  on the client devices at a training round, as shown in Figure~\ref{fig:devicelatency}. Compared with the on-device training latency that remained constant at different rounds, the communication latency showed high variability at each round, showing up to 3.91$\times$ increase over the mean communication latency. Our testbed experiment had higher variability than our previous simulated experiments, as the mean CV (Coefficient of Variance) of communication latency was 1.5$\times$ larger (0.59 over 0.40).}

\rev{Such variable communication latency affected each FL method differently. For baselines that use fixed deadlines (e.g., 1$T$), clients often failed to send their model updates due to long communication latency. For SmartPC (SPC) and Wait-for-All (WFA) baselines, the server had to wait for a prolonged duration when the clients were experiencing poor network connectivity. As the network conditions were not identical at different experiments, 
some baselines (FedAvg+2$T$, FedAvg+SPC, Prox+1$T$, and Prox+2$T$) yielded performance with huge variance ($\geq$ 0.20 in speedup). While client failures also negatively affected \putname{}, it achieved superior time-to-accuracy and accuracy performance over the baselines due to its adaptive deadline configuration with DDL-E measurement on the sampled clients at a training round.}



\rev{We suspect the variability of the communication latency would be higher in real deployments as users with mobility and unstable network conditions would be involved. Moreover, the impact of variable communication latency would be more significant when we train larger models with >100M parameters (e.g., BERT~\cite{devlin01}) in FL. As \putname{} is not designed to actively respond to the network connectivity changes of clients in real-time, we expect \putname{} could be improved further if the client-side network condition analysis system is integrated to accurately predict the round completion time of selected clients at each round. We leave this as future work.}
\section{Discussion}


\textbf{Local Epoch Training Policies.}
While \putname{} was originally designed for FL based on FedAvg~\cite{mcmahan01} that performs full data training per local epoch of client, recent studies such as Oort~\cite{lai01} propose to perform single batch training per local epoch. There are pros and cons in both approaches; single batch training offers more frequent global model update with shorter training time on clients, but this could lead to excessive communication overhead when training large models. Full data training exploits full statistical utility of client data per round, but offers less frequent global model update with longer round. While it is the model developer's role to determine the option, \putname{} is readily applicable and improves time-to-accuracy performance on both as we demonstrated in the evaluation.

\textbf{Effect of Forward Pass at a Client.}
\label{sec:latencytradeoff}
When \putname{} selects a clients' samples, it uses a list of sample loss that is maintained on a client from the beginning of FL, instead of performing a forward pass on the client data at each round. A sample loss at the maintained list is only updated whenever the sample is selected and trained by a client, which may result in containing outdated information. We conducted an experiment to understand the effect of forward pass on performance of \putname{}, which is shown in Figure~\ref{fig:forwardpass}. On FEMNIST dataset, integrating the forward pass with \putname{} resulted in slower training, which supports our design without the forward pass. This is because that the outdated loss values are generally larger than the newly-updated ones, which encourages \putname{} to select more diverse samples while prioritizing informative samples. 


\textbf{Robustness of Sample Selection.}
One of the possible limitation of \putname{} is that it might perform worse on FL tasks with noisy data, as noisy samples are highly likely to be selected by the sample selection module that prioritizes high loss. As we observed the performance improvement with \putname{} on five real-world user datasets which may already have certain noise level, we expect \putname{} would be helpful on most FL tasks. To improve further, we could systematically involve robust training approaches at centralized learning~\cite{shen01, roh01, song02} to actively deal with noisy data. This is part of our future work.

\rev{\textbf{Potential Bias of Sample Selection.}
While FedBalancer prioritizes more ``informative'' samples for training, it might integrate more samples from certain classes or sensitive groups than others, potentially leading to performance degradation on less sampled entities. To address this issue, we could integrate our sample selection strategy with other sample selection or reweighting approaches~\cite{roh01, yan01} that are designed to achieve unbiased model training. We leave this as future work.}

\section{Related Work}
\label{sec:relatedwork}

We survey closely related work with \putname{} other than the FL approaches on heterogeneous clients which we discussed earlier in Section~\ref{sec:motivation}.

\textbf{Sample Selection in Machine Learning.} There are several sample selection approaches in the field of machine learning research that could be arranged in threefold: (1) \textit{Curriculum Learning (CL)}~\cite{wu01, bengio01, graves01, hacohen01, huang01}: 
CL is a sample ordering technique which trains a network with easier samples in early training stage and gradually increase the difficulty to improve convergence speed and model generalization. 
However, applying it in FL is challenging as CL require a \textit{reference model} to determine the difficulty of samples, which hardly exists in FL scenarios. (2) \textit{Active Learning (AL)}~\cite{settles01, balcan01, gal01, kirsch01}: AL is a sample selection technique on unlabeled data, which interactively queries the user to label new data points that is likely to be more informative to the given task. Applying AL in FL could be non-trivial, as the training data is isolated and the labels are known but not shared externally from the clients.
(3) \textit{Importance Sampling}~\cite{katharopoulos01, alain01, loshchilov01, schaul01}: Being motivated by the fact that the importance of each training samples is different, researchers have proposed importance sampling techniques to accelerate the model training. 
While their idea could be brought to FL to prioritize samples during training, determining \textit{how many} and \textit{which} samples to use for each training round and \textit{when} to calculate the sample importance is yet unknown. 


\textbf{Sample Selection in FL.} Tuor et al.~\cite{tuor01} proposed a scheme that selects relevant clients' data to the given FL task, but it only selects the dataset before the FL starts. Moreover, it requires an example dataset, which is hardly applicable at FL scenarios where the client data distributions are usually unknown. 
\rev{Li et al.~\cite{li08} 
proposes how we can prioritize client training samples with higher importance in FL using \textit{gradient norm upper bound}~\cite{katharopoulos01}. However, their approach does not provide \textit{how many} samples should be selected per each round. While other methods such as FedSS~\cite{cai01} determines the amount of client training samples during FL, it does not specify which samples to select, simply adopting random sampling of the data. Moreover, combining FedSS with Li et al. is nontrivial, as FedSS assumes random sampling of the data. Unlike previous approaches, \putname{} is the first systematic framework that actively determines (1) \textit{how many} and (2) \textit{which} samples to select during FL to improve time-to-accuracy performance. We believe such design of \putname{} enables the use of client sample selection to improve time-to-accuracy performance.}

\textbf{Deadline Control in FL.} 
Determining an optimal deadline has been largely overlooked by previous approaches; only 
SmartPC determines a deadline to enable a specified proportion of the devices to complete a training round. \rev{It assumes that every client uses the same set of data for each round of FL. 
\putname{} on the other hand utilizes a new deadline control strategy for FL that enables high convergence speed where client training samples dynamically change during FL due to sample selection.}
\section{Conclusion}

We presented \putname{}, a systematic FL framework with sample selection for optimized training process. \putname{} actively selects the samples with high statistical utility through client-server coordination at each FL round without exposing private information of users. To further accelerate FL with our sample selection, we design adaptive deadline control strategy for \putname{} to predict the optimal deadline for each round with client sample selection. Our evaluation of on five real-world datasets from three different domains reveal that \putname{} achieves \rev{1.20$\sim$4.48$\times$} speedup over existing FL algorithms with different deadline configuration methods, while improving the model accuracy by \rev{1.1$\sim$5.0\%}. Our design of \putname{} is easily applicable on top of orthogonal FL methods, that we demonstrate the joint implementation of \putname{} with three existing FL algorithms and report the improved time-to-accuracy performance and model accuracy.

\begin{acks}

We thank anonymous reviewers and our shepherd, Lin Zhong, for their constructive suggestions.
This work was supported in part by the National Research Foundation of Korea (NRF) grant funded by the Korea government (MSIT)  (No.NRF-2020R1A2C1004062) and the Institute of Information \& communications Technology Planning \& evaluation (IITP) grant funded by the Korean government (MSIT) (No. 2021-0-00900).

\end{acks}

\bibliographystyle{ACM-Reference-Format}
\bibliography{main}


\begin{thebibliography}{79}


\ifx \showCODEN    \undefined \def \showCODEN     #1{\unskip}     \fi
\ifx \showDOI      \undefined \def \showDOI       #1{#1}\fi
\ifx \showISBNx    \undefined \def \showISBNx     #1{\unskip}     \fi
\ifx \showISBNxiii \undefined \def \showISBNxiii  #1{\unskip}     \fi
\ifx \showISSN     \undefined \def \showISSN      #1{\unskip}     \fi
\ifx \showLCCN     \undefined \def \showLCCN      #1{\unskip}     \fi
\ifx \shownote     \undefined \def \shownote      #1{#1}          \fi
\ifx \showarticletitle \undefined \def \showarticletitle #1{#1}   \fi
\ifx \showURL      \undefined \def \showURL       {\relax}        \fi
\providecommand\bibfield[2]{#2}
\providecommand\bibinfo[2]{#2}
\providecommand\natexlab[1]{#1}
\providecommand\showeprint[2][]{arXiv:#2}

\bibitem[\protect\citeauthoryear{??}{pro}{[n.d.]}]%
        {protocolbuffers}
 \bibinfo{year}{[n.d.]}\natexlab{}.
\newblock \bibinfo{title}{Protocol Buffers}.
\newblock
  \bibinfo{howpublished}{\url{https://developers.google.com/protocol-buffers/docs/overview}}.
\newblock
\newblock
\shownote{Accessed: 2022-05-17.}


\bibitem[\protect\citeauthoryear{??}{tfl}{[n.d.]}]%
        {tflite}
 \bibinfo{year}{[n.d.]}\natexlab{}.
\newblock \bibinfo{title}{TensorFlow Lite}.
\newblock \bibinfo{howpublished}{\url{https://www.tensorflow.org/lite}}.
\newblock
\newblock
\shownote{Accessed: 2022-05-17.}


\bibitem[\protect\citeauthoryear{Abadi, Chu, Goodfellow, McMahan, Mironov,
  Talwar, and Zhang}{Abadi et~al\mbox{.}}{2016}]%
        {abadi01}
\bibfield{author}{\bibinfo{person}{Martin Abadi}, \bibinfo{person}{Andy Chu},
  \bibinfo{person}{Ian Goodfellow}, \bibinfo{person}{H.~Brendan McMahan},
  \bibinfo{person}{Ilya Mironov}, \bibinfo{person}{Kunal Talwar}, {and}
  \bibinfo{person}{Li Zhang}.} \bibinfo{year}{2016}\natexlab{}.
\newblock \showarticletitle{Deep Learning with Differential Privacy}. In
  \bibinfo{booktitle}{\emph{Proceedings of the 2016 ACM SIGSAC Conference on
  Computer and Communications Security}} (Vienna, Austria)
  \emph{(\bibinfo{series}{CCS '16})}. \bibinfo{publisher}{Association for
  Computing Machinery}, \bibinfo{address}{New York, NY, USA},
  \bibinfo{pages}{308–318}.
\newblock
\showISBNx{9781450341394}
\urldef\tempurl%
\url{https://doi.org/10.1145/2976749.2978318}
\showDOI{\tempurl}


\bibitem[\protect\citeauthoryear{Abdelmoniem, Ho, Papageorgiou, Bilal, and
  Canini}{Abdelmoniem et~al\mbox{.}}{2021}]%
        {abdelmoniem01}
\bibfield{author}{\bibinfo{person}{Ahmed~M. Abdelmoniem},
  \bibinfo{person}{Chen{-}Yu Ho}, \bibinfo{person}{Pantelis Papageorgiou},
  \bibinfo{person}{Muhammad Bilal}, {and} \bibinfo{person}{Marco Canini}.}
  \bibinfo{year}{2021}\natexlab{}.
\newblock \showarticletitle{On the Impact of Device and Behavioral
  Heterogeneity in Federated Learning}.
\newblock \bibinfo{journal}{\emph{CoRR}}  \bibinfo{volume}{abs/2102.07500}
  (\bibinfo{year}{2021}).
\newblock
\showeprint[arXiv]{2102.07500}
\urldef\tempurl%
\url{https://arxiv.org/abs/2102.07500}
\showURL{%
\tempurl}


\bibitem[\protect\citeauthoryear{Alain, Lamb, Sankar, Courville, and
  Bengio}{Alain et~al\mbox{.}}{2015}]%
        {alain01}
\bibfield{author}{\bibinfo{person}{Guillaume Alain}, \bibinfo{person}{Alex
  Lamb}, \bibinfo{person}{Chinnadhurai Sankar}, \bibinfo{person}{Aaron
  Courville}, {and} \bibinfo{person}{Yoshua Bengio}.}
  \bibinfo{year}{2015}\natexlab{}.
\newblock \showarticletitle{Variance reduction in sgd by distributed importance
  sampling}.
\newblock \bibinfo{journal}{\emph{arXiv preprint arXiv:1511.06481}}
  (\bibinfo{year}{2015}).
\newblock


\bibitem[\protect\citeauthoryear{Albaseer, Ciftler, Abdallah, and
  Al-Fuqaha}{Albaseer et~al\mbox{.}}{2020}]%
        {albaseer01}
\bibfield{author}{\bibinfo{person}{Abdullatif Albaseer},
  \bibinfo{person}{Bekir~Sait Ciftler}, \bibinfo{person}{Mohamed Abdallah},
  {and} \bibinfo{person}{Ala Al-Fuqaha}.} \bibinfo{year}{2020}\natexlab{}.
\newblock \showarticletitle{Exploiting unlabeled data in smart cities using
  federated edge learning}. In \bibinfo{booktitle}{\emph{2020 International
  Wireless Communications and Mobile Computing (IWCMC)}}. IEEE,
  \bibinfo{pages}{1666--1671}.
\newblock


\bibitem[\protect\citeauthoryear{Anguita, Ghio, Oneto, Parra, Reyes-Ortiz,
  et~al\mbox{.}}{Anguita et~al\mbox{.}}{2013}]%
        {anguita01}
\bibfield{author}{\bibinfo{person}{Davide Anguita}, \bibinfo{person}{Alessandro
  Ghio}, \bibinfo{person}{Luca Oneto}, \bibinfo{person}{Xavier Parra},
  \bibinfo{person}{Jorge~Luis Reyes-Ortiz}, {et~al\mbox{.}}}
  \bibinfo{year}{2013}\natexlab{}.
\newblock \showarticletitle{A public domain dataset for human activity
  recognition using smartphones.}. In \bibinfo{booktitle}{\emph{21th European
  Symposium on Artificial Neural Networks, Computational Intelligence and
  Machine Learning, (ESANN'13)}}.
\newblock


\bibitem[\protect\citeauthoryear{Balcan, Beygelzimer, and Langford}{Balcan
  et~al\mbox{.}}{2009}]%
        {balcan01}
\bibfield{author}{\bibinfo{person}{Maria-Florina Balcan},
  \bibinfo{person}{Alina Beygelzimer}, {and} \bibinfo{person}{John Langford}.}
  \bibinfo{year}{2009}\natexlab{}.
\newblock \showarticletitle{Agnostic active learning}.
\newblock \bibinfo{journal}{\emph{J. Comput. System Sci.}}
  \bibinfo{volume}{75}, \bibinfo{number}{1} (\bibinfo{year}{2009}),
  \bibinfo{pages}{78--89}.
\newblock


\bibitem[\protect\citeauthoryear{Bengio, Louradour, Collobert, and
  Weston}{Bengio et~al\mbox{.}}{2009}]%
        {bengio01}
\bibfield{author}{\bibinfo{person}{Yoshua Bengio},
  \bibinfo{person}{J{\'e}r{\^o}me Louradour}, \bibinfo{person}{Ronan
  Collobert}, {and} \bibinfo{person}{Jason Weston}.}
  \bibinfo{year}{2009}\natexlab{}.
\newblock \showarticletitle{Curriculum learning}. In
  \bibinfo{booktitle}{\emph{Proceedings of the 26th annual international
  conference on machine learning}}. \bibinfo{pages}{41--48}.
\newblock


\bibitem[\protect\citeauthoryear{Beutel, Topal, Mathur, Qiu, Parcollet, and
  Lane}{Beutel et~al\mbox{.}}{2020}]%
        {beutel01}
\bibfield{author}{\bibinfo{person}{Daniel~J Beutel}, \bibinfo{person}{Taner
  Topal}, \bibinfo{person}{Akhil Mathur}, \bibinfo{person}{Xinchi Qiu},
  \bibinfo{person}{Titouan Parcollet}, {and} \bibinfo{person}{Nicholas~D
  Lane}.} \bibinfo{year}{2020}\natexlab{}.
\newblock \showarticletitle{Flower: A Friendly Federated Learning Research
  Framework}.
\newblock \bibinfo{journal}{\emph{arXiv preprint arXiv:2007.14390}}
  (\bibinfo{year}{2020}).
\newblock


\bibitem[\protect\citeauthoryear{Bonawitz, Eichner, Grieskamp, Huba, Ingerman,
  Ivanov, Kiddon, Kone{\v{c}}n{\`y}, Mazzocchi, McMahan,
  et~al\mbox{.}}{Bonawitz et~al\mbox{.}}{2019}]%
        {bonawitz01}
\bibfield{author}{\bibinfo{person}{Keith Bonawitz}, \bibinfo{person}{Hubert
  Eichner}, \bibinfo{person}{Wolfgang Grieskamp}, \bibinfo{person}{Dzmitry
  Huba}, \bibinfo{person}{Alex Ingerman}, \bibinfo{person}{Vladimir Ivanov},
  \bibinfo{person}{Chloe Kiddon}, \bibinfo{person}{Jakub Kone{\v{c}}n{\`y}},
  \bibinfo{person}{Stefano Mazzocchi}, \bibinfo{person}{H~Brendan McMahan},
  {et~al\mbox{.}}} \bibinfo{year}{2019}\natexlab{}.
\newblock \showarticletitle{Towards federated learning at scale: System
  design}.
\newblock \bibinfo{journal}{\emph{arXiv preprint arXiv:1902.01046}}
  (\bibinfo{year}{2019}).
\newblock


\bibitem[\protect\citeauthoryear{Cai, Lin, Zhang, and Yu}{Cai
  et~al\mbox{.}}{2020}]%
        {cai01}
\bibfield{author}{\bibinfo{person}{Lingshuang Cai}, \bibinfo{person}{Di Lin},
  \bibinfo{person}{Jiale Zhang}, {and} \bibinfo{person}{Shui Yu}.}
  \bibinfo{year}{2020}\natexlab{}.
\newblock \showarticletitle{Dynamic sample selection for federated learning
  with heterogeneous data in fog computing}. In \bibinfo{booktitle}{\emph{ICC
  2020-2020 IEEE International Conference on Communications (ICC)}}. IEEE,
  \bibinfo{pages}{1--6}.
\newblock


\bibitem[\protect\citeauthoryear{Caldas, Duddu, Wu, Li, Kone{\v{c}}n{\`y},
  McMahan, Smith, and Talwalkar}{Caldas et~al\mbox{.}}{2018}]%
        {caldas01}
\bibfield{author}{\bibinfo{person}{Sebastian Caldas}, \bibinfo{person}{Sai
  Meher~Karthik Duddu}, \bibinfo{person}{Peter Wu}, \bibinfo{person}{Tian Li},
  \bibinfo{person}{Jakub Kone{\v{c}}n{\`y}}, \bibinfo{person}{H~Brendan
  McMahan}, \bibinfo{person}{Virginia Smith}, {and} \bibinfo{person}{Ameet
  Talwalkar}.} \bibinfo{year}{2018}\natexlab{}.
\newblock \showarticletitle{Leaf: A benchmark for federated settings}.
\newblock \bibinfo{journal}{\emph{arXiv preprint arXiv:1812.01097}}
  (\bibinfo{year}{2018}).
\newblock


\bibitem[\protect\citeauthoryear{Cho, Gupta, Joshi, and Ya{\u{g}}an}{Cho
  et~al\mbox{.}}{2020a}]%
        {cho02}
\bibfield{author}{\bibinfo{person}{Yae~Jee Cho}, \bibinfo{person}{Samarth
  Gupta}, \bibinfo{person}{Gauri Joshi}, {and} \bibinfo{person}{Osman
  Ya{\u{g}}an}.} \bibinfo{year}{2020}\natexlab{a}.
\newblock \showarticletitle{Bandit-based Communication-Efficient Client
  Selection Strategies for Federated Learning}. In
  \bibinfo{booktitle}{\emph{2020 54th Asilomar Conference on Signals, Systems,
  and Computers}}. IEEE, \bibinfo{pages}{1066--1069}.
\newblock


\bibitem[\protect\citeauthoryear{Cho, Wang, and Joshi}{Cho
  et~al\mbox{.}}{2020b}]%
        {cho01}
\bibfield{author}{\bibinfo{person}{Yae~Jee Cho}, \bibinfo{person}{Jianyu Wang},
  {and} \bibinfo{person}{Gauri Joshi}.} \bibinfo{year}{2020}\natexlab{b}.
\newblock \showarticletitle{Client selection in federated learning: Convergence
  analysis and power-of-choice selection strategies}.
\newblock \bibinfo{journal}{\emph{arXiv preprint arXiv:2010.01243}}
  (\bibinfo{year}{2020}).
\newblock


\bibitem[\protect\citeauthoryear{Ciftler, Albaseer, Lasla, and
  Abdallah}{Ciftler et~al\mbox{.}}{2020}]%
        {ciftler01}
\bibfield{author}{\bibinfo{person}{Bekir~Sait Ciftler},
  \bibinfo{person}{Abdullatif Albaseer}, \bibinfo{person}{Noureddine Lasla},
  {and} \bibinfo{person}{Mohamed Abdallah}.} \bibinfo{year}{2020}\natexlab{}.
\newblock \showarticletitle{Federated learning for localization: A
  privacy-preserving crowdsourcing method}.
\newblock \bibinfo{journal}{\emph{arXiv preprint arXiv:2001.01911}}
  (\bibinfo{year}{2020}).
\newblock


\bibitem[\protect\citeauthoryear{Cohen, Afshar, Tapson, and Van~Schaik}{Cohen
  et~al\mbox{.}}{2017}]%
        {cohen01}
\bibfield{author}{\bibinfo{person}{Gregory Cohen}, \bibinfo{person}{Saeed
  Afshar}, \bibinfo{person}{Jonathan Tapson}, {and} \bibinfo{person}{Andre
  Van~Schaik}.} \bibinfo{year}{2017}\natexlab{}.
\newblock \showarticletitle{EMNIST: Extending MNIST to handwritten letters}. In
  \bibinfo{booktitle}{\emph{2017 International Joint Conference on Neural
  Networks (IJCNN)}}. IEEE, \bibinfo{pages}{2921--2926}.
\newblock


\bibitem[\protect\citeauthoryear{Dayan, Roth, Zhong, Harouni, Gentili, Abidin,
  Liu, Costa, Wood, Tsai, et~al\mbox{.}}{Dayan et~al\mbox{.}}{2021}]%
        {dayan01}
\bibfield{author}{\bibinfo{person}{Ittai Dayan}, \bibinfo{person}{Holger~R
  Roth}, \bibinfo{person}{Aoxiao Zhong}, \bibinfo{person}{Ahmed Harouni},
  \bibinfo{person}{Amilcare Gentili}, \bibinfo{person}{Anas~Z Abidin},
  \bibinfo{person}{Andrew Liu}, \bibinfo{person}{Anthony~Beardsworth Costa},
  \bibinfo{person}{Bradford~J Wood}, \bibinfo{person}{Chien-Sung Tsai},
  {et~al\mbox{.}}} \bibinfo{year}{2021}\natexlab{}.
\newblock \showarticletitle{Federated learning for predicting clinical outcomes
  in patients with COVID-19}.
\newblock \bibinfo{journal}{\emph{Nature medicine}} \bibinfo{volume}{27},
  \bibinfo{number}{10} (\bibinfo{year}{2021}), \bibinfo{pages}{1735--1743}.
\newblock


\bibitem[\protect\citeauthoryear{Devlin, Chang, Lee, and Toutanova}{Devlin
  et~al\mbox{.}}{2019}]%
        {devlin01}
\bibfield{author}{\bibinfo{person}{Jacob Devlin}, \bibinfo{person}{Ming-Wei
  Chang}, \bibinfo{person}{Kenton Lee}, {and} \bibinfo{person}{Kristina
  Toutanova}.} \bibinfo{year}{2019}\natexlab{}.
\newblock \showarticletitle{{BERT}: Pre-training of Deep Bidirectional
  Transformers for Language Understanding}. In
  \bibinfo{booktitle}{\emph{Proceedings of the 2019 Conference of the North
  {A}merican Chapter of the Association for Computational Linguistics: Human
  Language Technologies, Volume 1 (Long and Short Papers)}}.
  \bibinfo{publisher}{Association for Computational Linguistics},
  \bibinfo{address}{Minneapolis, Minnesota}, \bibinfo{pages}{4171--4186}.
\newblock
\urldef\tempurl%
\url{https://doi.org/10.18653/v1/N19-1423}
\showDOI{\tempurl}


\bibitem[\protect\citeauthoryear{Diao, Ding, and Tarokh}{Diao
  et~al\mbox{.}}{2021}]%
        {diao01}
\bibfield{author}{\bibinfo{person}{Enmao Diao}, \bibinfo{person}{Jie Ding},
  {and} \bibinfo{person}{Vahid Tarokh}.} \bibinfo{year}{2021}\natexlab{}.
\newblock \showarticletitle{HeteroFL: Computation and Communication Efficient
  Federated Learning for Heterogeneous Clients}. In
  \bibinfo{booktitle}{\emph{9th International Conference on Learning
  Representations, {ICLR} 2021, Virtual Event, Austria, May 3-7, 2021}}.
  \bibinfo{publisher}{OpenReview.net}.
\newblock
\urldef\tempurl%
\url{https://openreview.net/forum?id=TNkPBBYFkXg}
\showURL{%
\tempurl}


\bibitem[\protect\citeauthoryear{Dinh, Nguyen, Hoang, Vu, and Dutkiewicz}{Dinh
  et~al\mbox{.}}{2021}]%
        {dinh01}
\bibfield{author}{\bibinfo{person}{Thinh~Quang Dinh}, \bibinfo{person}{Diep~N
  Nguyen}, \bibinfo{person}{Dinh~Thai Hoang}, \bibinfo{person}{Pham~Tran Vu},
  {and} \bibinfo{person}{Eryk Dutkiewicz}.} \bibinfo{year}{2021}\natexlab{}.
\newblock \showarticletitle{In-network Computation for Large-scale Federated
  Learning over Wireless Edge Networks}.
\newblock \bibinfo{journal}{\emph{arXiv preprint arXiv:2109.10903}}
  (\bibinfo{year}{2021}).
\newblock


\bibitem[\protect\citeauthoryear{Ek, Portet, Lalanda, and Baez}{Ek
  et~al\mbox{.}}{2021}]%
        {ek01}
\bibfield{author}{\bibinfo{person}{Sannara Ek}, \bibinfo{person}{Fran{\c{c}}ois
  Portet}, \bibinfo{person}{Philippe Lalanda}, {and} \bibinfo{person}{German
  Eduardo~Vega Baez}.} \bibinfo{year}{2021}\natexlab{}.
\newblock \showarticletitle{Evaluating Federated Learning for human activity
  recognition}. In \bibinfo{booktitle}{\emph{Workshop AI for Internet of
  Things, in conjunction with IJCAI-PRICAI 2020}}.
\newblock


\bibitem[\protect\citeauthoryear{Fallah, Mokhtari, and Ozdaglar}{Fallah
  et~al\mbox{.}}{2020}]%
        {fallah01}
\bibfield{author}{\bibinfo{person}{Alireza Fallah}, \bibinfo{person}{Aryan
  Mokhtari}, {and} \bibinfo{person}{Asuman~E Ozdaglar}.}
  \bibinfo{year}{2020}\natexlab{}.
\newblock \showarticletitle{Personalized Federated Learning with Theoretical
  Guarantees: A Model-Agnostic Meta-Learning Approach.}. In
  \bibinfo{booktitle}{\emph{NeurIPS}}.
\newblock


\bibitem[\protect\citeauthoryear{Feng, Rong, Sun, Guo, and Li}{Feng
  et~al\mbox{.}}{2020}]%
        {feng01}
\bibfield{author}{\bibinfo{person}{Jie Feng}, \bibinfo{person}{Can Rong},
  \bibinfo{person}{Funing Sun}, \bibinfo{person}{Diansheng Guo}, {and}
  \bibinfo{person}{Yong Li}.} \bibinfo{year}{2020}\natexlab{}.
\newblock \showarticletitle{PMF: A privacy-preserving human mobility prediction
  framework via federated learning}.
\newblock \bibinfo{journal}{\emph{Proceedings of the ACM on Interactive,
  Mobile, Wearable and Ubiquitous Technologies}} \bibinfo{volume}{4},
  \bibinfo{number}{1} (\bibinfo{year}{2020}), \bibinfo{pages}{1--21}.
\newblock


\bibitem[\protect\citeauthoryear{Gal, Islam, and Ghahramani}{Gal
  et~al\mbox{.}}{2017}]%
        {gal01}
\bibfield{author}{\bibinfo{person}{Yarin Gal}, \bibinfo{person}{Riashat Islam},
  {and} \bibinfo{person}{Zoubin Ghahramani}.} \bibinfo{year}{2017}\natexlab{}.
\newblock \showarticletitle{Deep bayesian active learning with image data}. In
  \bibinfo{booktitle}{\emph{International Conference on Machine Learning}}.
  PMLR, \bibinfo{pages}{1183--1192}.
\newblock


\bibitem[\protect\citeauthoryear{Graves, Bellemare, Menick, Munos, and
  Kavukcuoglu}{Graves et~al\mbox{.}}{2017}]%
        {graves01}
\bibfield{author}{\bibinfo{person}{Alex Graves}, \bibinfo{person}{Marc~G
  Bellemare}, \bibinfo{person}{Jacob Menick}, \bibinfo{person}{Remi Munos},
  {and} \bibinfo{person}{Koray Kavukcuoglu}.} \bibinfo{year}{2017}\natexlab{}.
\newblock \showarticletitle{Automated curriculum learning for neural networks}.
  In \bibinfo{booktitle}{\emph{international conference on machine learning}}.
  PMLR, \bibinfo{pages}{1311--1320}.
\newblock


\bibitem[\protect\citeauthoryear{Hacohen and Weinshall}{Hacohen and
  Weinshall}{2019}]%
        {hacohen01}
\bibfield{author}{\bibinfo{person}{Guy Hacohen} {and} \bibinfo{person}{Daphna
  Weinshall}.} \bibinfo{year}{2019}\natexlab{}.
\newblock \showarticletitle{On the power of curriculum learning in training
  deep networks}. In \bibinfo{booktitle}{\emph{International Conference on
  Machine Learning}}. PMLR, \bibinfo{pages}{2535--2544}.
\newblock


\bibitem[\protect\citeauthoryear{Horv{\'a}th, Laskaridis, Almeida, Leontiadis,
  Venieris, and Lane}{Horv{\'a}th et~al\mbox{.}}{2021}]%
        {horvath01}
\bibfield{author}{\bibinfo{person}{Samuel Horv{\'a}th},
  \bibinfo{person}{Stefanos Laskaridis}, \bibinfo{person}{Mario Almeida},
  \bibinfo{person}{Ilias Leontiadis}, \bibinfo{person}{Stylianos Venieris},
  {and} \bibinfo{person}{Nicholas~Donald Lane}.}
  \bibinfo{year}{2021}\natexlab{}.
\newblock \showarticletitle{Fj{ORD}: Fair and Accurate Federated Learning under
  heterogeneous targets with Ordered Dropout}. In
  \bibinfo{booktitle}{\emph{Thirty-Fifth Conference on Neural Information
  Processing Systems}}.
\newblock
\urldef\tempurl%
\url{https://openreview.net/forum?id=4fLr7H5D_eT}
\showURL{%
\tempurl}


\bibitem[\protect\citeauthoryear{Huang, Wang, Tai, Liu, Shen, Li, Li, and
  Huang}{Huang et~al\mbox{.}}{2020}]%
        {huang01}
\bibfield{author}{\bibinfo{person}{Yuge Huang}, \bibinfo{person}{Yuhan Wang},
  \bibinfo{person}{Ying Tai}, \bibinfo{person}{Xiaoming Liu},
  \bibinfo{person}{Pengcheng Shen}, \bibinfo{person}{Shaoxin Li},
  \bibinfo{person}{Jilin Li}, {and} \bibinfo{person}{Feiyue Huang}.}
  \bibinfo{year}{2020}\natexlab{}.
\newblock \showarticletitle{Curricularface: adaptive curriculum learning loss
  for deep face recognition}. In \bibinfo{booktitle}{\emph{proceedings of the
  IEEE/CVF conference on computer vision and pattern recognition}}.
  \bibinfo{pages}{5901--5910}.
\newblock


\bibitem[\protect\citeauthoryear{Jiang, Kantarci, Oktug, and Soyata}{Jiang
  et~al\mbox{.}}{2020}]%
        {jiang01}
\bibfield{author}{\bibinfo{person}{Ji~Chu Jiang}, \bibinfo{person}{Burak
  Kantarci}, \bibinfo{person}{Sema Oktug}, {and} \bibinfo{person}{Tolga
  Soyata}.} \bibinfo{year}{2020}\natexlab{}.
\newblock \showarticletitle{Federated learning in smart city sensing:
  Challenges and opportunities}.
\newblock \bibinfo{journal}{\emph{Sensors}} \bibinfo{volume}{20},
  \bibinfo{number}{21} (\bibinfo{year}{2020}), \bibinfo{pages}{6230}.
\newblock


\bibitem[\protect\citeauthoryear{Jiang, Kone{\v{c}}n{\`y}, Rush, and
  Kannan}{Jiang et~al\mbox{.}}{2019}]%
        {jiang02}
\bibfield{author}{\bibinfo{person}{Yihan Jiang}, \bibinfo{person}{Jakub
  Kone{\v{c}}n{\`y}}, \bibinfo{person}{Keith Rush}, {and}
  \bibinfo{person}{Sreeram Kannan}.} \bibinfo{year}{2019}\natexlab{}.
\newblock \showarticletitle{Improving federated learning personalization via
  model agnostic meta learning}.
\newblock \bibinfo{journal}{\emph{arXiv preprint arXiv:1909.12488}}
  (\bibinfo{year}{2019}).
\newblock


\bibitem[\protect\citeauthoryear{Jim{\'e}nez-S{\'a}nchez, Tardy, Ballester,
  Mateus, and Piella}{Jim{\'e}nez-S{\'a}nchez et~al\mbox{.}}{2021}]%
        {jimenez01}
\bibfield{author}{\bibinfo{person}{Amelia Jim{\'e}nez-S{\'a}nchez},
  \bibinfo{person}{Mickael Tardy}, \bibinfo{person}{Miguel A~Gonz{\'a}lez
  Ballester}, \bibinfo{person}{Diana Mateus}, {and} \bibinfo{person}{Gemma
  Piella}.} \bibinfo{year}{2021}\natexlab{}.
\newblock \showarticletitle{Memory-aware curriculum federated learning for
  breast cancer classification}.
\newblock \bibinfo{journal}{\emph{arXiv preprint arXiv:2107.02504}}
  (\bibinfo{year}{2021}).
\newblock


\bibitem[\protect\citeauthoryear{Kairouz, McMahan, Avent, Bellet, Bennis,
  Bhagoji, Bonawitz, et~al\mbox{.}}{Kairouz et~al\mbox{.}}{2019}]%
        {kairouz01}
\bibfield{author}{\bibinfo{person}{Peter Kairouz}, \bibinfo{person}{H.~Brendan
  McMahan}, \bibinfo{person}{Brendan Avent}, \bibinfo{person}{Aurélien
  Bellet}, \bibinfo{person}{Mehdi Bennis}, \bibinfo{person}{Arjun~Nitin
  Bhagoji}, \bibinfo{person}{K.~A. Bonawitz}, {et~al\mbox{.}}}
  \bibinfo{year}{2019}\natexlab{}.
\newblock \showarticletitle{Advances and Open Problems in Federated Learning}.
\newblock
\urldef\tempurl%
\url{https://arxiv.org/abs/1912.04977}
\showURL{%
\tempurl}


\bibitem[\protect\citeauthoryear{Katharopoulos and Fleuret}{Katharopoulos and
  Fleuret}{2018}]%
        {katharopoulos01}
\bibfield{author}{\bibinfo{person}{Angelos Katharopoulos} {and}
  \bibinfo{person}{Fran{\c{c}}ois Fleuret}.} \bibinfo{year}{2018}\natexlab{}.
\newblock \showarticletitle{Not all samples are created equal: Deep learning
  with importance sampling}. In \bibinfo{booktitle}{\emph{International
  conference on machine learning}}. PMLR, \bibinfo{pages}{2525--2534}.
\newblock


\bibitem[\protect\citeauthoryear{Kirkpatrick, Pascanu, Rabinowitz, Veness,
  Desjardins, Rusu, Milan, Quan, Ramalho, Grabska-Barwinska,
  et~al\mbox{.}}{Kirkpatrick et~al\mbox{.}}{2017}]%
        {kirkpatrick01}
\bibfield{author}{\bibinfo{person}{James Kirkpatrick}, \bibinfo{person}{Razvan
  Pascanu}, \bibinfo{person}{Neil Rabinowitz}, \bibinfo{person}{Joel Veness},
  \bibinfo{person}{Guillaume Desjardins}, \bibinfo{person}{Andrei~A Rusu},
  \bibinfo{person}{Kieran Milan}, \bibinfo{person}{John Quan},
  \bibinfo{person}{Tiago Ramalho}, \bibinfo{person}{Agnieszka
  Grabska-Barwinska}, {et~al\mbox{.}}} \bibinfo{year}{2017}\natexlab{}.
\newblock \showarticletitle{Overcoming catastrophic forgetting in neural
  networks}.
\newblock \bibinfo{journal}{\emph{Proceedings of the national academy of
  sciences}} \bibinfo{volume}{114}, \bibinfo{number}{13}
  (\bibinfo{year}{2017}), \bibinfo{pages}{3521--3526}.
\newblock


\bibitem[\protect\citeauthoryear{Kirsch, Van~Amersfoort, and Gal}{Kirsch
  et~al\mbox{.}}{2019}]%
        {kirsch01}
\bibfield{author}{\bibinfo{person}{Andreas Kirsch}, \bibinfo{person}{Joost
  Van~Amersfoort}, {and} \bibinfo{person}{Yarin Gal}.}
  \bibinfo{year}{2019}\natexlab{}.
\newblock \showarticletitle{Batchbald: Efficient and diverse batch acquisition
  for deep bayesian active learning}.
\newblock \bibinfo{journal}{\emph{Advances in neural information processing
  systems}}  \bibinfo{volume}{32} (\bibinfo{year}{2019}),
  \bibinfo{pages}{7026--7037}.
\newblock


\bibitem[\protect\citeauthoryear{Kone{\v{c}}n{\`y}, McMahan, Yu, Richt{\'a}rik,
  Suresh, and Bacon}{Kone{\v{c}}n{\`y} et~al\mbox{.}}{2016}]%
        {konevcny01}
\bibfield{author}{\bibinfo{person}{Jakub Kone{\v{c}}n{\`y}},
  \bibinfo{person}{H~Brendan McMahan}, \bibinfo{person}{Felix~X Yu},
  \bibinfo{person}{Peter Richt{\'a}rik}, \bibinfo{person}{Ananda~Theertha
  Suresh}, {and} \bibinfo{person}{Dave Bacon}.}
  \bibinfo{year}{2016}\natexlab{}.
\newblock \showarticletitle{Federated learning: Strategies for improving
  communication efficiency}.
\newblock \bibinfo{journal}{\emph{arXiv preprint arXiv:1610.05492}}
  (\bibinfo{year}{2016}).
\newblock


\bibitem[\protect\citeauthoryear{Lai, Zhu, Madhyastha, and Chowdhury}{Lai
  et~al\mbox{.}}{2021}]%
        {lai01}
\bibfield{author}{\bibinfo{person}{Fan Lai}, \bibinfo{person}{Xiangfeng Zhu},
  \bibinfo{person}{Harsha~V. Madhyastha}, {and} \bibinfo{person}{Mosharaf
  Chowdhury}.} \bibinfo{year}{2021}\natexlab{}.
\newblock \showarticletitle{Oort: Efficient Federated Learning via Guided
  Participant Selection}. In \bibinfo{booktitle}{\emph{15th {USENIX} Symposium
  on Operating Systems Design and Implementation ({OSDI}'21)}}.
  \bibinfo{pages}{19--35}.
\newblock


\bibitem[\protect\citeauthoryear{Le, Lei, Mu, Zhang, Zeng, and Liao}{Le
  et~al\mbox{.}}{2021}]%
        {le01}
\bibfield{author}{\bibinfo{person}{Junqing Le}, \bibinfo{person}{Xinyu Lei},
  \bibinfo{person}{Nankun Mu}, \bibinfo{person}{Hengrun Zhang},
  \bibinfo{person}{Kai Zeng}, {and} \bibinfo{person}{Xiaofeng Liao}.}
  \bibinfo{year}{2021}\natexlab{}.
\newblock \showarticletitle{Federated Continuous Learning With Broad Network
  Architecture}.
\newblock \bibinfo{journal}{\emph{IEEE Transactions on Cybernetics}}
  \bibinfo{volume}{51}, \bibinfo{number}{8} (\bibinfo{year}{2021}),
  \bibinfo{pages}{3874--3888}.
\newblock


\bibitem[\protect\citeauthoryear{Li, Sun, Li, Pu, Li, and Chen}{Li
  et~al\mbox{.}}{2021b}]%
        {li07}
\bibfield{author}{\bibinfo{person}{Ang Li}, \bibinfo{person}{Jingwei Sun},
  \bibinfo{person}{Pengcheng Li}, \bibinfo{person}{Yu Pu}, \bibinfo{person}{Hai
  Li}, {and} \bibinfo{person}{Yiran Chen}.} \bibinfo{year}{2021}\natexlab{b}.
\newblock \showarticletitle{Hermes: an efficient federated learning framework
  for heterogeneous mobile clients}. In \bibinfo{booktitle}{\emph{Proceedings
  of the 27th Annual International Conference on Mobile Computing and
  Networking}}. \bibinfo{pages}{420--437}.
\newblock


\bibitem[\protect\citeauthoryear{Li, Sun, Zeng, Zhang, Li, and Chen}{Li
  et~al\mbox{.}}{2021c}]%
        {li06}
\bibfield{author}{\bibinfo{person}{Ang Li}, \bibinfo{person}{Jingwei Sun},
  \bibinfo{person}{Xiao Zeng}, \bibinfo{person}{Mi Zhang}, \bibinfo{person}{Hai
  Li}, {and} \bibinfo{person}{Yiran Chen}.} \bibinfo{year}{2021}\natexlab{c}.
\newblock \showarticletitle{FedMask: Joint Computation and
  Communication-Efficient Personalized Federated Learning via Heterogeneous
  Masking}. In \bibinfo{booktitle}{\emph{Proceedings of the 19th ACM Conference
  on Embedded Networked Sensor Systems}}. \bibinfo{pages}{42--55}.
\newblock


\bibitem[\protect\citeauthoryear{Li, Zhang, Tan, Qin, Wang, and Li}{Li
  et~al\mbox{.}}{2021d}]%
        {li08}
\bibfield{author}{\bibinfo{person}{Anran Li}, \bibinfo{person}{Lan Zhang},
  \bibinfo{person}{Juntao Tan}, \bibinfo{person}{Yaxuan Qin},
  \bibinfo{person}{Junhao Wang}, {and} \bibinfo{person}{Xiang-Yang Li}.}
  \bibinfo{year}{2021}\natexlab{d}.
\newblock \showarticletitle{Sample-level Data Selection for Federated
  Learning}. In \bibinfo{booktitle}{\emph{IEEE INFOCOM 2021-IEEE Conference on
  Computer Communications}}. IEEE, \bibinfo{pages}{1--10}.
\newblock


\bibitem[\protect\citeauthoryear{Li, Xiong, Guo, Wang, and Xu}{Li
  et~al\mbox{.}}{2019b}]%
        {li04}
\bibfield{author}{\bibinfo{person}{Li Li}, \bibinfo{person}{Haoyi Xiong},
  \bibinfo{person}{Zhishan Guo}, \bibinfo{person}{Jun Wang}, {and}
  \bibinfo{person}{Cheng-Zhong Xu}.} \bibinfo{year}{2019}\natexlab{b}.
\newblock \showarticletitle{Smartpc: Hierarchical pace control in real-time
  federated learning system}. In \bibinfo{booktitle}{\emph{2019 IEEE Real-Time
  Systems Symposium (RTSS)}}. IEEE, \bibinfo{pages}{406--418}.
\newblock


\bibitem[\protect\citeauthoryear{Li, Diao, Chen, and He}{Li
  et~al\mbox{.}}{2021a}]%
        {li03}
\bibfield{author}{\bibinfo{person}{Qinbin Li}, \bibinfo{person}{Yiqun Diao},
  \bibinfo{person}{Quan Chen}, {and} \bibinfo{person}{Bingsheng He}.}
  \bibinfo{year}{2021}\natexlab{a}.
\newblock \showarticletitle{Federated Learning on Non-IID Data Silos: An
  Experimental Study}.
\newblock \bibinfo{journal}{\emph{arXiv preprint arXiv:2102.02079}}
  (\bibinfo{year}{2021}).
\newblock


\bibitem[\protect\citeauthoryear{Li, Sahu, Zaheer, Sanjabi, Talwalkar, and
  Smith}{Li et~al\mbox{.}}{2020a}]%
        {li01}
\bibfield{author}{\bibinfo{person}{Tian Li}, \bibinfo{person}{Anit~Kumar Sahu},
  \bibinfo{person}{Manzil Zaheer}, \bibinfo{person}{Maziar Sanjabi},
  \bibinfo{person}{Ameet Talwalkar}, {and} \bibinfo{person}{Virginia Smith}.}
  \bibinfo{year}{2020}\natexlab{a}.
\newblock \showarticletitle{Federated Optimization in Heterogeneous Networks}.
  In \bibinfo{booktitle}{\emph{Proceedings of Machine Learning and Systems
  2020, MLSys 2020, Austin, TX, USA, March 2-4, 2020}},
  \bibfield{editor}{\bibinfo{person}{Inderjit~S. Dhillon},
  \bibinfo{person}{Dimitris~S. Papailiopoulos}, {and} \bibinfo{person}{Vivienne
  Sze}} (Eds.). \bibinfo{publisher}{mlsys.org}.
\newblock
\urldef\tempurl%
\url{https://proceedings.mlsys.org/book/316.pdf}
\showURL{%
\tempurl}


\bibitem[\protect\citeauthoryear{Li, Sanjabi, Beirami, and Smith}{Li
  et~al\mbox{.}}{2020b}]%
        {li05}
\bibfield{author}{\bibinfo{person}{Tian Li}, \bibinfo{person}{Maziar Sanjabi},
  \bibinfo{person}{Ahmad Beirami}, {and} \bibinfo{person}{Virginia Smith}.}
  \bibinfo{year}{2020}\natexlab{b}.
\newblock \showarticletitle{Fair Resource Allocation in Federated Learning}. In
  \bibinfo{booktitle}{\emph{8th International Conference on Learning
  Representations, {ICLR} 2020, Addis Ababa, Ethiopia, April 26-30, 2020}}.
  \bibinfo{publisher}{OpenReview.net}.
\newblock
\urldef\tempurl%
\url{https://openreview.net/forum?id=ByexElSYDr}
\showURL{%
\tempurl}


\bibitem[\protect\citeauthoryear{Li, Milletar{\`\i}, Xu, Rieke, Hancox, Zhu,
  Baust, Cheng, Ourselin, Cardoso, et~al\mbox{.}}{Li et~al\mbox{.}}{2019a}]%
        {li02}
\bibfield{author}{\bibinfo{person}{Wenqi Li}, \bibinfo{person}{Fausto
  Milletar{\`\i}}, \bibinfo{person}{Daguang Xu}, \bibinfo{person}{Nicola
  Rieke}, \bibinfo{person}{Jonny Hancox}, \bibinfo{person}{Wentao Zhu},
  \bibinfo{person}{Maximilian Baust}, \bibinfo{person}{Yan Cheng},
  \bibinfo{person}{S{\'e}bastien Ourselin}, \bibinfo{person}{M~Jorge Cardoso},
  {et~al\mbox{.}}} \bibinfo{year}{2019}\natexlab{a}.
\newblock \showarticletitle{Privacy-preserving federated brain tumour
  segmentation}. In \bibinfo{booktitle}{\emph{International workshop on machine
  learning in medical imaging}}. Springer, \bibinfo{pages}{133--141}.
\newblock


\bibitem[\protect\citeauthoryear{Liu, Cai, Zhang, Li, Wang, Li, Guo, and
  Chen}{Liu et~al\mbox{.}}{2021}]%
        {liu2021distfl}
\bibfield{author}{\bibinfo{person}{Bingyan Liu}, \bibinfo{person}{Yifeng Cai},
  \bibinfo{person}{Ziqi Zhang}, \bibinfo{person}{Yuanchun Li},
  \bibinfo{person}{Leye Wang}, \bibinfo{person}{Ding Li}, \bibinfo{person}{Yao
  Guo}, {and} \bibinfo{person}{Xiangqun Chen}.}
  \bibinfo{year}{2021}\natexlab{}.
\newblock \showarticletitle{DistFL: Distribution-aware Federated Learning for
  Mobile Scenarios}.
\newblock \bibinfo{journal}{\emph{Proceedings of the ACM on Interactive,
  Mobile, Wearable and Ubiquitous Technologies}} \bibinfo{volume}{5},
  \bibinfo{number}{4} (\bibinfo{year}{2021}), \bibinfo{pages}{1--26}.
\newblock


\bibitem[\protect\citeauthoryear{Liu, Li, Liu, Guo, and Chen}{Liu
  et~al\mbox{.}}{2020}]%
        {liu2020pmc}
\bibfield{author}{\bibinfo{person}{Bingyan Liu}, \bibinfo{person}{Yuanchun Li},
  \bibinfo{person}{Yunxin Liu}, \bibinfo{person}{Yao Guo}, {and}
  \bibinfo{person}{Xiangqun Chen}.} \bibinfo{year}{2020}\natexlab{}.
\newblock \showarticletitle{Pmc: A privacy-preserving deep learning model
  customization framework for edge computing}.
\newblock \bibinfo{journal}{\emph{Proceedings of the ACM on Interactive,
  Mobile, Wearable and Ubiquitous Technologies}} \bibinfo{volume}{4},
  \bibinfo{number}{4} (\bibinfo{year}{2020}), \bibinfo{pages}{1--25}.
\newblock


\bibitem[\protect\citeauthoryear{Liu, Luo, Wang, and Tang}{Liu
  et~al\mbox{.}}{2015}]%
        {liu01}
\bibfield{author}{\bibinfo{person}{Ziwei Liu}, \bibinfo{person}{Ping Luo},
  \bibinfo{person}{Xiaogang Wang}, {and} \bibinfo{person}{Xiaoou Tang}.}
  \bibinfo{year}{2015}\natexlab{}.
\newblock \showarticletitle{Deep Learning Face Attributes in the Wild}. In
  \bibinfo{booktitle}{\emph{Proceedings of International Conference on Computer
  Vision (ICCV)}}.
\newblock


\bibitem[\protect\citeauthoryear{Loshchilov and Hutter}{Loshchilov and
  Hutter}{2015}]%
        {loshchilov01}
\bibfield{author}{\bibinfo{person}{Ilya Loshchilov} {and}
  \bibinfo{person}{Frank Hutter}.} \bibinfo{year}{2015}\natexlab{}.
\newblock \showarticletitle{Online batch selection for faster training of
  neural networks}.
\newblock \bibinfo{journal}{\emph{arXiv preprint arXiv:1511.06343}}
  (\bibinfo{year}{2015}).
\newblock


\bibitem[\protect\citeauthoryear{McMahan, Moore, Ramage, Hampson, and
  y~Arcas}{McMahan et~al\mbox{.}}{2017}]%
        {mcmahan01}
\bibfield{author}{\bibinfo{person}{Brendan McMahan}, \bibinfo{person}{Eider
  Moore}, \bibinfo{person}{Daniel Ramage}, \bibinfo{person}{Seth Hampson},
  {and} \bibinfo{person}{Blaise~Ag{\"{u}}era y Arcas}.}
  \bibinfo{year}{2017}\natexlab{}.
\newblock \showarticletitle{Communication-Efficient Learning of Deep Networks
  from Decentralized Data}. In \bibinfo{booktitle}{\emph{Proceedings of the
  20th International Conference on Artificial Intelligence and Statistics,
  {AISTATS} 2017, 20-22 April 2017, Fort Lauderdale, FL, {USA}}}
  \emph{(\bibinfo{series}{Proceedings of Machine Learning Research},
  Vol.~\bibinfo{volume}{54})}, \bibfield{editor}{\bibinfo{person}{Aarti Singh}
  {and} \bibinfo{person}{Xiaojin~(Jerry) Zhu}} (Eds.).
  \bibinfo{publisher}{{PMLR}}, \bibinfo{pages}{1273--1282}.
\newblock
\urldef\tempurl%
\url{http://proceedings.mlr.press/v54/mcmahan17a.html}
\showURL{%
\tempurl}


\bibitem[\protect\citeauthoryear{McMahan, Ramage, Talwar, and Zhang}{McMahan
  et~al\mbox{.}}{2018}]%
        {mcmahan02}
\bibfield{author}{\bibinfo{person}{H.~Brendan McMahan}, \bibinfo{person}{Daniel
  Ramage}, \bibinfo{person}{Kunal Talwar}, {and} \bibinfo{person}{Li Zhang}.}
  \bibinfo{year}{2018}\natexlab{}.
\newblock \showarticletitle{Learning Differentially Private Recurrent Language
  Models}. In \bibinfo{booktitle}{\emph{6th International Conference on
  Learning Representations, {ICLR} 2018, Vancouver, BC, Canada, April 30 - May
  3, 2018, Conference Track Proceedings}}. \bibinfo{publisher}{OpenReview.net}.
\newblock
\urldef\tempurl%
\url{https://openreview.net/forum?id=BJ0hF1Z0b}
\showURL{%
\tempurl}


\bibitem[\protect\citeauthoryear{Nishio and Yonetani}{Nishio and
  Yonetani}{2019}]%
        {nishio01}
\bibfield{author}{\bibinfo{person}{Takayuki Nishio} {and} \bibinfo{person}{Ryo
  Yonetani}.} \bibinfo{year}{2019}\natexlab{}.
\newblock \showarticletitle{Client selection for federated learning with
  heterogeneous resources in mobile edge}. In \bibinfo{booktitle}{\emph{ICC
  2019-2019 IEEE International Conference on Communications (ICC)}}. IEEE,
  \bibinfo{pages}{1--7}.
\newblock


\bibitem[\protect\citeauthoryear{Niu, Wu, Tang, Hua, Jia, Lv, Wu, and Chen}{Niu
  et~al\mbox{.}}{2020}]%
        {niu01}
\bibfield{author}{\bibinfo{person}{Chaoyue Niu}, \bibinfo{person}{Fan Wu},
  \bibinfo{person}{Shaojie Tang}, \bibinfo{person}{Lifeng Hua},
  \bibinfo{person}{Rongfei Jia}, \bibinfo{person}{Chengfei Lv},
  \bibinfo{person}{Zhihua Wu}, {and} \bibinfo{person}{Guihai Chen}.}
  \bibinfo{year}{2020}\natexlab{}.
\newblock \showarticletitle{Billion-scale federated learning on mobile clients:
  a submodel design with tunable privacy}. In
  \bibinfo{booktitle}{\emph{Proceedings of the 26th Annual International
  Conference on Mobile Computing and Networking}}. \bibinfo{pages}{1--14}.
\newblock


\bibitem[\protect\citeauthoryear{Ouyang, Xie, Zhou, Huang, and Xing}{Ouyang
  et~al\mbox{.}}{2021}]%
        {ouyang01}
\bibfield{author}{\bibinfo{person}{Xiaomin Ouyang}, \bibinfo{person}{Zhiyuan
  Xie}, \bibinfo{person}{Jiayu Zhou}, \bibinfo{person}{Jianwei Huang}, {and}
  \bibinfo{person}{Guoliang Xing}.} \bibinfo{year}{2021}\natexlab{}.
\newblock \showarticletitle{ClusterFL: A Similarity-Aware Federated Learning
  System for Human Activity Recognition}. In
  \bibinfo{booktitle}{\emph{Proceedings of the 19th Annual International
  Conference on Mobile Systems, Applications, and Services}} (Virtual Event,
  Wisconsin) \emph{(\bibinfo{series}{MobiSys '21})}.
  \bibinfo{publisher}{Association for Computing Machinery},
  \bibinfo{address}{New York, NY, USA}, \bibinfo{pages}{54–66}.
\newblock
\showISBNx{9781450384438}
\urldef\tempurl%
\url{https://doi.org/10.1145/3458864.3467681}
\showDOI{\tempurl}


\bibitem[\protect\citeauthoryear{Pilla}{Pilla}{2021}]%
        {pilla01}
\bibfield{author}{\bibinfo{person}{La{\'e}rcio~Lima Pilla}.}
  \bibinfo{year}{2021}\natexlab{}.
\newblock \showarticletitle{Optimal Task Assignment for Heterogeneous Federated
  Learning Devices}. In \bibinfo{booktitle}{\emph{2021 IEEE International
  Parallel and Distributed Processing Symposium (IPDPS)}}. IEEE,
  \bibinfo{pages}{661--670}.
\newblock


\bibitem[\protect\citeauthoryear{Pillutla, Roulet, Kakade, and
  Harchaoui}{Pillutla et~al\mbox{.}}{2018}]%
        {pillutla01}
\bibfield{author}{\bibinfo{person}{Venkata~K Pillutla},
  \bibinfo{person}{Vincent Roulet}, \bibinfo{person}{Sham~M Kakade}, {and}
  \bibinfo{person}{Zaid Harchaoui}.} \bibinfo{year}{2018}\natexlab{}.
\newblock \showarticletitle{A smoother way to train structured prediction
  models}.
\newblock \bibinfo{journal}{\emph{Advances in Neural Information Processing
  Systems 31}} (\bibinfo{year}{2018}).
\newblock


\bibitem[\protect\citeauthoryear{Rios and Sahinidis}{Rios and
  Sahinidis}{2013}]%
        {rios01}
\bibfield{author}{\bibinfo{person}{Luis~Miguel Rios} {and}
  \bibinfo{person}{Nikolaos~V Sahinidis}.} \bibinfo{year}{2013}\natexlab{}.
\newblock \showarticletitle{Derivative-free optimization: a review of
  algorithms and comparison of software implementations}.
\newblock \bibinfo{journal}{\emph{Journal of Global Optimization}}
  \bibinfo{volume}{56}, \bibinfo{number}{3} (\bibinfo{year}{2013}),
  \bibinfo{pages}{1247--1293}.
\newblock


\bibitem[\protect\citeauthoryear{Roh, Lee, Whang, and Suh}{Roh
  et~al\mbox{.}}{2021}]%
        {roh01}
\bibfield{author}{\bibinfo{person}{Yuji Roh}, \bibinfo{person}{Kangwook Lee},
  \bibinfo{person}{Steven~Euijong Whang}, {and} \bibinfo{person}{Changho Suh}.}
  \bibinfo{year}{2021}\natexlab{}.
\newblock \showarticletitle{Sample Selection for Fair and Robust Training}. In
  \bibinfo{booktitle}{\emph{NeurIPS}}.
\newblock


\bibitem[\protect\citeauthoryear{Schaul, Quan, Antonoglou, and Silver}{Schaul
  et~al\mbox{.}}{2016}]%
        {schaul01}
\bibfield{author}{\bibinfo{person}{Tom Schaul}, \bibinfo{person}{John Quan},
  \bibinfo{person}{Ioannis Antonoglou}, {and} \bibinfo{person}{David Silver}.}
  \bibinfo{year}{2016}\natexlab{}.
\newblock \showarticletitle{Prioritized Experience Replay}. In
  \bibinfo{booktitle}{\emph{4th International Conference on Learning
  Representations, {ICLR} 2016, San Juan, Puerto Rico, May 2-4, 2016,
  Conference Track Proceedings}}, \bibfield{editor}{\bibinfo{person}{Yoshua
  Bengio} {and} \bibinfo{person}{Yann LeCun}} (Eds.).
\newblock
\urldef\tempurl%
\url{http://arxiv.org/abs/1511.05952}
\showURL{%
\tempurl}


\bibitem[\protect\citeauthoryear{Settles}{Settles}{2009}]%
        {settles01}
\bibfield{author}{\bibinfo{person}{Burr Settles}.}
  \bibinfo{year}{2009}\natexlab{}.
\newblock \showarticletitle{Active learning literature survey}.
\newblock  (\bibinfo{year}{2009}).
\newblock


\bibitem[\protect\citeauthoryear{Shakespeare}{Shakespeare}{2014}]%
        {shakespeare01}
\bibfield{author}{\bibinfo{person}{William Shakespeare}.}
  \bibinfo{year}{2014}\natexlab{}.
\newblock \bibinfo{booktitle}{\emph{The complete works of William
  Shakespeare}}.
\newblock \bibinfo{publisher}{Race Point Publishing}.
\newblock


\bibitem[\protect\citeauthoryear{Shen and Sanghavi}{Shen and Sanghavi}{2019}]%
        {shen01}
\bibfield{author}{\bibinfo{person}{Yanyao Shen} {and} \bibinfo{person}{Sujay
  Sanghavi}.} \bibinfo{year}{2019}\natexlab{}.
\newblock \showarticletitle{Learning with Bad Training Data via Iterative
  Trimmed Loss Minimization}. In \bibinfo{booktitle}{\emph{ICML}}.
  \bibinfo{pages}{5739–5748}.
\newblock


\bibitem[\protect\citeauthoryear{Song, Kim, and Lee}{Song
  et~al\mbox{.}}{2019}]%
        {song02}
\bibfield{author}{\bibinfo{person}{Hwanjun Song}, \bibinfo{person}{Minseok
  Kim}, {and} \bibinfo{person}{Jae-Gil Lee}.} \bibinfo{year}{2019}\natexlab{}.
\newblock \showarticletitle{Selfie: Refurbishing unclean samples for robust
  deep learning}. In \bibinfo{booktitle}{\emph{ICML}}.
  \bibinfo{pages}{5907–5915}.
\newblock


\bibitem[\protect\citeauthoryear{Song, Kim, Park, Shin, and Lee}{Song
  et~al\mbox{.}}{2020}]%
        {song01}
\bibfield{author}{\bibinfo{person}{Hwanjun Song}, \bibinfo{person}{Minseok
  Kim}, \bibinfo{person}{Dongmin Park}, \bibinfo{person}{Yooju Shin}, {and}
  \bibinfo{person}{Jae-Gil Lee}.} \bibinfo{year}{2020}\natexlab{}.
\newblock \showarticletitle{Learning from noisy labels with deep neural
  networks: A survey}.
\newblock \bibinfo{journal}{\emph{arXiv preprint arXiv:2007.08199}}
  (\bibinfo{year}{2020}).
\newblock


\bibitem[\protect\citeauthoryear{Sozinov, Vlassov, and Girdzijauskas}{Sozinov
  et~al\mbox{.}}{2018}]%
        {sozinov01}
\bibfield{author}{\bibinfo{person}{Konstantin Sozinov},
  \bibinfo{person}{Vladimir Vlassov}, {and} \bibinfo{person}{Sarunas
  Girdzijauskas}.} \bibinfo{year}{2018}\natexlab{}.
\newblock \showarticletitle{Human activity recognition using federated
  learning}. In \bibinfo{booktitle}{\emph{2018 IEEE Intl Conf on Parallel \&
  Distributed Processing with Applications, Ubiquitous Computing \&
  Communications, Big Data \& Cloud Computing, Social Computing \& Networking,
  Sustainable Computing \& Communications
  (ISPA/IUCC/BDCloud/SocialCom/SustainCom)}}. IEEE,
  \bibinfo{pages}{1103--1111}.
\newblock


\bibitem[\protect\citeauthoryear{Tu, Ouyang, Zhou, He, and Xing}{Tu
  et~al\mbox{.}}{2021}]%
        {tu01}
\bibfield{author}{\bibinfo{person}{Linlin Tu}, \bibinfo{person}{Xiaomin
  Ouyang}, \bibinfo{person}{Jiayu Zhou}, \bibinfo{person}{Yuze He}, {and}
  \bibinfo{person}{Guoliang Xing}.} \bibinfo{year}{2021}\natexlab{}.
\newblock \showarticletitle{FedDL: Federated Learning via Dynamic Layer Sharing
  for Human Activity Recognition}. In \bibinfo{booktitle}{\emph{Proceedings of
  the 19th ACM Conference on Embedded Networked Sensor Systems}}.
  \bibinfo{pages}{15--28}.
\newblock


\bibitem[\protect\citeauthoryear{Tuor, Wang, Ko, Liu, and Leung}{Tuor
  et~al\mbox{.}}{2020}]%
        {tuor01}
\bibfield{author}{\bibinfo{person}{Tiffany Tuor}, \bibinfo{person}{Shiqiang
  Wang}, \bibinfo{person}{Bong{-}Jun Ko}, \bibinfo{person}{Changchang Liu},
  {and} \bibinfo{person}{Kin~K. Leung}.} \bibinfo{year}{2020}\natexlab{}.
\newblock \showarticletitle{Overcoming Noisy and Irrelevant Data in Federated
  Learning}. In \bibinfo{booktitle}{\emph{25th International Conference on
  Pattern Recognition, {ICPR} 2020, Virtual Event / Milan, Italy, January
  10-15, 2021}}. \bibinfo{publisher}{{IEEE}}, \bibinfo{pages}{5020--5027}.
\newblock
\urldef\tempurl%
\url{https://doi.org/10.1109/ICPR48806.2021.9412599}
\showDOI{\tempurl}


\bibitem[\protect\citeauthoryear{Wang, Wei, and Zhou}{Wang
  et~al\mbox{.}}{2020a}]%
        {wang02}
\bibfield{author}{\bibinfo{person}{Cong Wang}, \bibinfo{person}{Xin Wei}, {and}
  \bibinfo{person}{Pengzhan Zhou}.} \bibinfo{year}{2020}\natexlab{a}.
\newblock \showarticletitle{Optimize scheduling of federated learning on
  battery-powered mobile devices}. In \bibinfo{booktitle}{\emph{2020 IEEE
  International Parallel and Distributed Processing Symposium (IPDPS)}}. IEEE,
  \bibinfo{pages}{212--221}.
\newblock


\bibitem[\protect\citeauthoryear{Wang, Yurochkin, Sun, Papailiopoulos, and
  Khazaeni}{Wang et~al\mbox{.}}{2020b}]%
        {wang01}
\bibfield{author}{\bibinfo{person}{Hongyi Wang}, \bibinfo{person}{Mikhail
  Yurochkin}, \bibinfo{person}{Yuekai Sun}, \bibinfo{person}{Dimitris~S.
  Papailiopoulos}, {and} \bibinfo{person}{Yasaman Khazaeni}.}
  \bibinfo{year}{2020}\natexlab{b}.
\newblock \showarticletitle{Federated Learning with Matched Averaging}. In
  \bibinfo{booktitle}{\emph{8th International Conference on Learning
  Representations, {ICLR} 2020, Addis Ababa, Ethiopia, April 26-30, 2020}}.
  \bibinfo{publisher}{OpenReview.net}.
\newblock
\urldef\tempurl%
\url{https://openreview.net/forum?id=BkluqlSFDS}
\showURL{%
\tempurl}


\bibitem[\protect\citeauthoryear{Wei, Li, Ding, Ma, Yang, Farokhi, Jin, Quek,
  and Poor}{Wei et~al\mbox{.}}{2020}]%
        {wei01}
\bibfield{author}{\bibinfo{person}{Kang Wei}, \bibinfo{person}{Jun Li},
  \bibinfo{person}{Ming Ding}, \bibinfo{person}{Chuan Ma},
  \bibinfo{person}{Howard~H Yang}, \bibinfo{person}{Farhad Farokhi},
  \bibinfo{person}{Shi Jin}, \bibinfo{person}{Tony~QS Quek}, {and}
  \bibinfo{person}{H~Vincent Poor}.} \bibinfo{year}{2020}\natexlab{}.
\newblock \showarticletitle{Federated learning with differential privacy:
  Algorithms and performance analysis}.
\newblock \bibinfo{journal}{\emph{IEEE Transactions on Information Forensics
  and Security}}  \bibinfo{volume}{15} (\bibinfo{year}{2020}),
  \bibinfo{pages}{3454--3469}.
\newblock


\bibitem[\protect\citeauthoryear{Wu, Chen, Zhou, and Zhang}{Wu
  et~al\mbox{.}}{2020}]%
        {wu01}
\bibfield{author}{\bibinfo{person}{Qiong Wu}, \bibinfo{person}{Xu Chen},
  \bibinfo{person}{Zhi Zhou}, {and} \bibinfo{person}{Junshan Zhang}.}
  \bibinfo{year}{2020}\natexlab{}.
\newblock \showarticletitle{FedHome: Cloud-Edge based Personalized Federated
  Learning for In-Home Health Monitoring}.
\newblock \bibinfo{journal}{\emph{IEEE Transactions on Mobile Computing}}
  (\bibinfo{year}{2020}).
\newblock


\bibitem[\protect\citeauthoryear{Yan, Seto, and Apostoloff}{Yan
  et~al\mbox{.}}{2022}]%
        {yan01}
\bibfield{author}{\bibinfo{person}{Bobby Yan}, \bibinfo{person}{Skyler Seto},
  {and} \bibinfo{person}{Nicholas Apostoloff}.}
  \bibinfo{year}{2022}\natexlab{}.
\newblock \showarticletitle{FORML: Learning to Reweight Data for Fairness}.
\newblock \bibinfo{journal}{\emph{arXiv preprint arXiv:2202.01719}}
  (\bibinfo{year}{2022}).
\newblock


\bibitem[\protect\citeauthoryear{Yang, Wang, Xu, Chen, Bian, Liu, and Liu}{Yang
  et~al\mbox{.}}{2021}]%
        {yang01}
\bibfield{author}{\bibinfo{person}{Chengxu Yang}, \bibinfo{person}{Qipeng
  Wang}, \bibinfo{person}{Mengwei Xu}, \bibinfo{person}{Zhenpeng Chen},
  \bibinfo{person}{Kaigui Bian}, \bibinfo{person}{Yunxin Liu}, {and}
  \bibinfo{person}{Xuanzhe Liu}.} \bibinfo{year}{2021}\natexlab{}.
\newblock \showarticletitle{Characterizing Impacts of Heterogeneity in
  Federated Learning upon Large-Scale Smartphone Data}. In
  \bibinfo{booktitle}{\emph{Proceedings of the Web Conference 2021}}.
  \bibinfo{pages}{935--946}.
\newblock


\bibitem[\protect\citeauthoryear{Yang, Lu, and Ren}{Yang et~al\mbox{.}}{2020}]%
        {yang03}
\bibfield{author}{\bibinfo{person}{Luting Yang}, \bibinfo{person}{Bingqian Lu},
  {and} \bibinfo{person}{Shaolei Ren}.} \bibinfo{year}{2020}\natexlab{}.
\newblock \showarticletitle{On the Latency Variability of Deep Neural Networks
  for Mobile Inference}. In \bibinfo{booktitle}{\emph{Workshop on Hot Topics in
  Edge Computing (1-page Poster)}}.
\newblock


\bibitem[\protect\citeauthoryear{Yang, Andrew, Eichner, Sun, Li, Kong, Ramage,
  and Beaufays}{Yang et~al\mbox{.}}{2018}]%
        {yang02}
\bibfield{author}{\bibinfo{person}{Timothy Yang}, \bibinfo{person}{Galen
  Andrew}, \bibinfo{person}{Hubert Eichner}, \bibinfo{person}{Haicheng Sun},
  \bibinfo{person}{Wei Li}, \bibinfo{person}{Nicholas Kong},
  \bibinfo{person}{Daniel Ramage}, {and} \bibinfo{person}{Fran{\c{c}}oise
  Beaufays}.} \bibinfo{year}{2018}\natexlab{}.
\newblock \showarticletitle{Applied federated learning: Improving google
  keyboard query suggestions}.
\newblock \bibinfo{journal}{\emph{arXiv preprint arXiv:1812.02903}}
  (\bibinfo{year}{2018}).
\newblock


\bibitem[\protect\citeauthoryear{Yoon, Madaan, Yang, and Hwang}{Yoon
  et~al\mbox{.}}{2021}]%
        {yoon01}
\bibfield{author}{\bibinfo{person}{Jaehong Yoon}, \bibinfo{person}{Divyam
  Madaan}, \bibinfo{person}{Eunho Yang}, {and} \bibinfo{person}{Sung~Ju
  Hwang}.} \bibinfo{year}{2021}\natexlab{}.
\newblock \showarticletitle{Online Coreset Selection for Rehearsal-based
  Continual Learning}.
\newblock \bibinfo{journal}{\emph{arXiv preprint arXiv:2106.01085}}
  (\bibinfo{year}{2021}).
\newblock


\bibitem[\protect\citeauthoryear{Zhao, Li, Lai, Suda, Civin, and Chandra}{Zhao
  et~al\mbox{.}}{2018}]%
        {zhao01}
\bibfield{author}{\bibinfo{person}{Yue Zhao}, \bibinfo{person}{Meng Li},
  \bibinfo{person}{Liangzhen Lai}, \bibinfo{person}{Naveen Suda},
  \bibinfo{person}{Damon Civin}, {and} \bibinfo{person}{Vikas Chandra}.}
  \bibinfo{year}{2018}\natexlab{}.
\newblock \showarticletitle{Federated learning with non-iid data}.
\newblock \bibinfo{journal}{\emph{arXiv preprint arXiv:1806.00582}}
  (\bibinfo{year}{2018}).
\newblock


\end{thebibliography}

\end{document}